\documentclass{article} 
\usepackage{nips12submit_e,times}
\usepackage{amssymb}
\usepackage{amsmath}
\usepackage{graphicx}
\usepackage{float}
\usepackage{wrapfig}
\usepackage[export]{adjustbox}

\title{Linear-Nonlinear-Poisson Neuron Networks Perform\\Bayesian Inference On Boltzmann Machines}

\author{
Louis Yuanlong Shao
\\
Department of Computer Science \& Engineering\\
The Ohio State University\\
\texttt{shaoyu@cse.ohio-state.edu}
}

\nipsfinalcopy 

\begin{document}

\maketitle

\begin{abstract}
One conjecture in both deep learning and classical connectionist viewpoint is that the biological brain implements certain kinds of deep networks as its back-end. However, to our knowledge, a detailed correspondence has not yet been set up, which is important if we want to bridge between neuroscience and machine learning. Recent researches emphasized the biological plausibility of Linear-Nonlinear-Poisson (LNP) neuron model. We show that with neurally plausible settings, the whole network is capable of representing any Boltzmann machine and performing a semi-stochastic Bayesian inference algorithm lying between Gibbs sampling and variational inference.
\end{abstract}

\section{Introduction}


Classical connectionist viewpoint has long been inspired from how the brain works, such as the invention of ``perceptron''~\cite{RosenblattPerceptron}, and the ``parallel distributed processing'' approach~\cite{PDPBook}. Modern deep learning and unsupervised feature learning also assume connections between biological brain and certain kinds of deep networks, either probabilistic or not. For example, properties of visual area V2 are found to be comparable to those on the sparse autoencoder networks~\cite{LeeSparseAutoencoder}; the sparse coding learning algorithm~\cite{OlshausenSparseCoding} is originated directly from neuroscience observations; also psychological phenomenon such as end-stopping is observed in sparse coding experiments~\cite{LeeSparseCoding}.

In addition, there are architectural similarity between neuroscience and machine learning. For example, neuronal spike propagation may correspond to prediction or inference tasks in a learned model; synaptic plasticity may correspond to parameter estimation given the network structure; neuroplasticity may correspond to learning the network structure together with inventing new hidden units in the network; axonal path finding in the help of guidance signals may be regarded as structural priors making learning network structure easier.

For this reason, it may be helpful again to refer to how the brain works when dealing with problems that are currently puzzling machine learning researchers. For example, training deep networks was made practical mostly after the breakthrough of deep learning starting from 2006 ({\em e.g.}~\cite{Hinton06Fast}\cite{Hinton06DimRed}\cite{Bengio07Greedy}\cite{Andrew12Building}\cite{Lee09Conv}, etc.). The success may lie in a safe way of adapting the network structure, such that the parameter estimation part can be effective enough to reach non-trivial local optima. However, since the brain is not exactly layer-wise, we may wonder is there any other way of choosing the network structure along with parameter estimation? If we want to transfer neuroscience knowledge for answering questions as such and build bridge between these two areas, we want to first figure out what exactly is the deep network that the brain is representing.

A good starting point of answering this may be to look at single neuron computation. Through how they transfer information among them, we can reverse engineer both the representation and the prediction / inference algorithm, which is the main purpose of this paper.


In theoretical neuroscience, single neuron models are divided into different levels of detail, (see~\cite{HerzModelingSingleNeuron} for a five level survey). Detailed models can be quite realistic and few simplification assumptions hold. Yet effective simplification holds in different extent. For example, integrate-and-fire models capture essential properties of Hodgkin-Huxley model~\cite{GerstnerSpikingNeuronModels}, and its variant exponential integrate-and-fire neuron models~\cite{ExpLIF} are found to be capable of reproducing spike timing of several types of cortical neurons. The community has also been investigating on the Linear-Nonlinear-Poisson (LNP) models for long~\cite{ChichilniskyLNP}\cite{Simoncelli04LNP}\cite{PaniMLCascade}. A recent study in~\cite{Spiking2Linear} analyzed how LNP models effectively reproduce firing rates in more realistic neurons. Based on these results, we start from formally presenting the LNP model and then turn to presenting a semi-stochastic inference algorithm on Boltzmann machines, which is derived from combining Gibbs sampling and variational inference. We then make a detailed matching between LNP model and the inference algorithm. Since the semi-stochastic inference algorithm has not been explored in the learning community, we also show some experiments illustrating its computational property, including its stochastic convergence and its similarity to variational inference.

\section{Brief Review on Neural Plausibility of LNP Model}
\label{sec:neuron}



In this section, we briefly review the Linear-Nonlinear-Poisson neuron model and its neural plausibility, focusing on what modeling options we have to fit it with useful inference algorithms.

The LNP neuron model~\cite{ChichilniskyLNP}\cite{Simoncelli04LNP}\cite{PaniMLCascade} formalizes the spike train generated by each neuron as a non-homogeneous Poisson point process, whose rate function over time depends on the input spike trains from its presynaptic neurons, with certain form of short-term memory in the dependence.
\begin{equation}
\label{eqn:LNPnew}\begin{gathered}
  {\text{Synaptic Transmission: }\text{  }\text{  }\text{  }}{I_{ij}}\left( \tau  \right) = {W_{ij}} \cdot \left( {{X_j} * \alpha } \right)\left( \tau  \right) \text{  }\text{  }\text{  }\text{  }\text{  }\text{  }\text{  }\text{  }\text{  }\text{  }\text{  }\text{  }\text{  }\text{  }\text{  }\text{  }\text{  }\text{  }\text{  }\text{  }\text{  }\text{  }\text{  }\text{  }\text{  }\text{  }\text{  }\text{(a)}\hfill \\
  {\text{Dendritic Integration: }\text{  }\text{  }\text{  }\text{  }\text{  }}{I_i}\left( \tau  \right) = {\text{Multi - Linear}}\left( {{{\left\{ {{I_{ij}}\left( \tau  \right)} \right\}}_{j \in M\left( i \right)}}} \right) \text{  }\text{  }\text{  }\text{  }\text{  }\text{  }\text{(b)}\hfill \\
  {\text{Spike Generation: }\text{  }\text{  }\text{  }\text{  }\text{  }\text{  }\text{  }\text{  }\text{  }\text{  }}\left\{ \begin{gathered}
  {\lambda _i}\left( \tau  \right) = \sigma \left( {\left( {{I_i} * D} \right)\left( \tau  \right)} \right) \hfill \text{  }\text{  }\text{  }\text{  }\text{  }\text{  }\text{(c)}\\
  {X_i}\left( \tau  \right)\sim{\text{PoissonPointProcess}}\left( {{\lambda _i}\left( \tau  \right)} \right) \hfill \text{  }\text{  }\text{  }\text{  }\text{  }\text{  }\text{  }\text{  }\text{(d)}\\
\end{gathered}  \right. \hfill \\
\end{gathered}
\end{equation}
In all these equations, $\tau$ is the continuous time index, whose unit is millisecond. Subscript $i$ is the index of the neuron in question, $M\left(i\right)$ is the index set of presynaptic neurons of neuron $i$. In Eq.~(\ref{eqn:LNPnew})(d), ${X_i}\left( \tau \right) = \sum\nolimits_f {\delta ( {\tau - \tau_i^{\left( f \right)}} )}$ represents the spike train, with spikes as Dirac functions located at time steps $\tau_i^{\left(f\right)}$. $\lambda_i\left(\tau\right)$ is the  rate function of the Poisson point process. Eq.~(\ref{eqn:LNPnew})(a) represents the postsynaptic current at a particular synapse with efficacy $W_{ij}$. $\alpha$ is a certain non-negative function with certain time course. Evidence for ``Poisson-like'' statistics in cortical neurons can be found in~\cite{TolhurstPoisson}\cite{MaPopulationCode}, which implies that spike counts over a certain interval may be Poisson distributed.

Eq.~(\ref{eqn:LNPnew})(b) is the dendritic summation. There are different viewpoints on whether linearity holds true. Due to phenomenon like mutual inhibition on different dendritic shafts, it is commonly known that nonlinear summation could happen (see~\cite{SegevDenTell}\cite{SegevDendriticProc} for example), yet studies from~\cite{ArayaDendriticLinearize} show that when dendritic spines are presented, linear summation is true, and this happens frequently in excitatory mammalian cortex. In addition, a survey on neuron models in \cite{HerzModelingSingleNeuron} summarized different computations that can be performed in dendritic processing, two notable examples are multiplication and summation. We include these two and generalizes them into the ``Multi-Linear'' function in Eq.~(\ref{eqn:LNPnew})(b), {\em i.e.}, it is linear with respect to each of its arguments individually. Here we use the set notation $\left\{\cdot\right\}_{j\in M\left(i\right)}$ to denote the set of arguments. What is special in a ``Multi-Linear'' function is that, when taking its expectation with respect to a certain argument, the expectation goes in and the resulted formula simply replaces the argument by its mean. This is useful for mean-field algorithms.

After dendritic summation, relationship between the summed current and the firing rate can be well researched by {\em in-vitro} neuroscience experiments where input current can be externally controlled. Various frequency-current function can also be derived from popular spike generation models, such as Leaky Integrate-and-Fire model (see~\cite{GerstnerSpikingNeuronModels} for an overview), Exponential Integrate-And-Fire Model~\cite{ExpLIF}, Spike Response Model~\cite{GerstnerSRM}, or more realistic models such as Hodgkin-Huxley Model~\cite{HodgkinHuxley}. In particular, \cite{ChichilniskyLNP}\cite{Simoncelli04LNP}\cite{PaniMLCascade} formulated the LNP simplification through a nonlinear mapping followed by a Poisson point process random sampling. The analytical study with simulation in~\cite{Spiking2Linear} show that the simplification is effective. We put this in Eq.~(\ref{eqn:LNPnew})(c).

\begin{wrapfigure}{r}{5cm} 
\centering
\includegraphics[width=5cm]{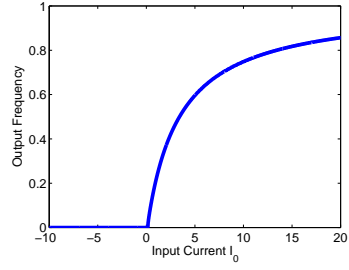}
\caption{Frequency-current curve of leaky integrate-and-fire neurons for a particular set of parameters. More from different neuron models can be found in~\cite{GerstnerSpikingNeuronModels}.}
\label{fig:gain}
\end{wrapfigure}

Since neural spikes has refractory period with at least 1ms, firing rate will not exceed 1000Hz and there won't be more than one spike happening in 1ms, so $\lambda_i\left(\tau\right) \in \left[0, 1\right]$. In practice, firing rate is often even lower. In addition, discrete time steps in the resolution of 1ms per each step seems quite fine-grained for most perceptual tasks (\emph{e.g.}, visual recognition emerges in 75-80ms~\cite{VanrullenVisualResponseTime}). For these reasons, discretizing time into 1ms per each step may be reasonable. By LeCam's theorem~\cite{LecamPoisson}, the discrete counterpart of nonhomogeneous Poisson point process is the nonhomogeneous Bernoulli process, and the Bernoulli probability at each discrete time step is approximated by the integration of rate function at that time period. We use $t\in \mathbb{Z}$ to denote the discrete time steps, whose unit is millisecond. We use $\phi_i^{\left(t\right)}\in \left[0, 1\right]$ to denote the Bernoulli probability and $x_i^{\left(t\right)} \in \left\{0, 1\right\}$ to denote the spike indicator at time $t$, 
\begin{equation}\label{eqn:defphi}
x_i^{\left( t \right)}\sim{\text{Bernoulli}}\left( {\phi _i^{\left( t \right)}} \right),{\text{  }\text{  }\text{  }}\phi _i^{\left( t \right)} \triangleq \int_t^{t + 1} {{\lambda _i}\left( \tau \right)d\tau}
\end{equation}
The corresponding discrete form of Eq.~(\ref{eqn:LNPnew})(a,b,c) is by substituting $X_i\left(\tau\right)$ with $x_i^{\left(t\right)}$ and substituting $\alpha$, $D$ functions with their discrete form similar to the definition of $\phi_i^{\left(t\right)}$ in Eq.~(\ref{eqn:defphi}). There are two ways to proceed with the formulation. In case $D$ function is the Dirac function located at 0, or is close to that, the convolution with $D$ can be ignored, the discrete counterpart of Eq.~(\ref{eqn:LNPnew})(a,b,c) are
\begin{equation}\label{eqn:mlnl}
\begin{gathered}
  I_{ij}^{\left( t \right)} = {W_{ij}} \cdot \sum\nolimits_{k = 1}^{K_{\alpha}} {\alpha \left( k \right) \cdot x_i^{\left( {t - k} \right)}}  \hfill \\
  \phi _i^{\left( t \right)} = \sigma \left( {{\text{Multi-Linear}}\left( {{{\left\{ {I_{ij}^{\left( t \right)}} \right\}}_{j \in M\left( i \right)}}} \right)} \right) \hfill \\
\end{gathered}
\end{equation}
Or if $D$ is non-trivial, but the Multi-Linear function in Eq.~(\ref{eqn:LNPnew})(b) reduces to a linear summation, by the associativity of convolution, we can let $\epsilon = \alpha * D$, and the discrete counterpart of Eq.~(\ref{eqn:LNPnew})(a,b,c) reduces to the following, assuming finite time course of the function $\epsilon$.
\begin{equation}\label{eqn:calcphi}
\phi _i^{\left( t \right)} = \sigma \left( {\sum\nolimits_{j \in M\left( i \right)} {{W_{ij}} \cdot \sum\nolimits_{k = 1}^K {\epsilon \left( k \right) \cdot x_i^{\left( {t - k} \right)}} } } \right)
\end{equation}
In~\cite{Spiking2Linear}, $D$ functions resulted from spiking neuron models are not Dirac functions, but are very close. In the latter part of this paper, we will focus on the case when linear summation is true (hence Eq.~(\ref{eqn:calcphi}) holds) although multi-linearity in Eq.~(\ref{eqn:mlnl}) may generalize our formulation to high-order Boltzmann machines. Another observation is that in Eq.~(\ref{eqn:calcphi}), if $D$ is exponential function (true in \cite{Spiking2Linear}) and $\alpha$ is exponential function too (true in \cite{GerstnerSpikingNeuronModels}), and assuming both functions have the same scale parameter, $\epsilon$ becomes the classic $\alpha$-function~\cite{Rall1967} used in postsynaptic modeling.

Another issue is whether linear summation in Eq.~(\ref{eqn:calcphi}) or multi-linear integration in Eq.~(\ref{eqn:mlnl}) contains a constant bias term $b_i$. In~\cite{GerstnerVariational}, there is such a bias term in the spiking response model. This will make our latter part easier, but in our derivation later, we focus on the case when there is no such bias. It will be trivial then to turn to the case where bias exists.

To sum up, the LNP neuron model we are going to deal with in the latter part is as the following, assuming $\mathcal{H}$ is the index set of neurons whose activity is not bound to observations (external stimuli).
\begin{equation}\label{eqn:neuron}
{\text{For all }}i \in \mathcal{H} , t \in {\mathbb{Z}^ + },{\rm{  }}\left\{ \begin{array}{l}
x_i^{\left( t \right)}\sim{\rm{Bernoulli}}\left( {\phi _i^{\left( t \right)}} \right)\\
\phi _i^{\left( t \right)} = \sigma \left( {\sum\nolimits_{j \in M\left( i \right)} {{W_{ij}}\cdot\sum\nolimits_{k = 1}^K {\epsilon \left( k \right)\cdot x_i^{\left( {t - k} \right)}} } } \right)
\end{array} \right.
\end{equation}
$\sigma$ function can be fit from data or from realistic neuron models, and has some relationship with the frequency-current curve. According to~\cite{GerstnerSpikingNeuronModels}\cite{Spiking2Linear}, such nonlinearity in different spiking neuron models may be non-decreasing, close to zero towards the left, increase almost linearly in a certain interval. Also, since firing rate has upper bounds, $\sigma$ won't increase unboundedly. For this reason, $\sigma$ may be considered as a modification of sigmoid function to meet certain demands of inference. An example of frequency-current curve is shown in Fig.~\ref{fig:gain}. See also~\cite{Spiking2Linear} for another type of nonlinearity. In Fig.~\ref{fig:gain}, parameters of the leaky integrate-and-fire model are set as ${\tau _m} = 20{\text{ ms, }}{\Delta ^{{\text{abs}}}} = 1{\text{ ms, }}R = 1{\text{ m}}\Omega {\text{, }}\upsilon  = 1{\text{ mV}}$ (with notations following~\cite{GerstnerSpikingNeuronModels}).

In the biological brain, all neurons are computing in parallel. Thus we want to investigate on the case when Eq.~(\ref{eqn:neuron}) is executed in parallel, vectors ${{\mathbf{x}}^{\left( t \right)}} = ( {x_i^{\left( t \right)}} )_{i = 1}^n$ and ${{\mathbf{\phi}}^{\left( t \right)}} = ( {\phi _i^{\left( t \right)}} )_{i = 1}^n$ over time forms a Markov chain of order $K$. The latter part focuses on showing its stochastic inference nature.


\section{Semi-Stochastic Inference on Boltzmann Machines}
\label{sec:mylabel1}

In this section, first we compare Gibbs sampling and variational inference on Boltzmann machines, emphasizing their architectural similarity, then we present the semi-stochastic inference algorithm by combining them. Some notations we use in this section will overlap with the last section. This is intended since we want to link the quantities in both sections.

Let ${\rm {\bf Y}}\in \left\{ {A,B} \right\}^{n}$ be a collection of random variables. Each of them has two possible values $A$ and $B$ ({\em e.g.}, $A=0$ and $B=1$ in the binary case). Let $\mathcal{V},\mathcal{H}$ be a partition of the index set $\mathcal{I}=\left\{ {1,...,n} \right\}$. We denote $Y_{\mathcal{V}} $ as the visible variables, $Y_{\mathcal{H}} $ as the hidden variables. We use lower case of $Y$ to denote its values, {\em e.g.}, $y_i$, $y_\mathcal{H}$ or $\mathbf{y}$. In this paper we use the Boltzmann machines with softmax units~\cite{RuslanSoftmax},
\begin{equation}
\label{eq4}\small
p\left( {\bf{y}} \right) = \frac{1}{Z}\exp \left( {\frac{1}{2} \cdot \sum\limits_{i,j:i \ne j;u,v \in \left\{ {A,B} \right\}} {\left[\kern-0.15em\left[ {{y_i} = u}
 \right]\kern-0.15em\right] \cdot \left[\kern-0.15em\left[ {{y_i} = v}
 \right]\kern-0.15em\right] \cdot {V_{iu,jv}}}  - \sum\limits_{i \in {\cal I},u \in \left\{ {A,B} \right\}} {\left[\kern-0.15em\left[ {{y_i} = u}
 \right]\kern-0.15em\right] \cdot {c_{iu}}} } \right)
\end{equation}
Here we denote $\left[\kern-0.15em\left[\cdot\right]\kern-0.15em\right]$ as the binary indicator. $\left[\kern-0.15em\left[\cdot\right]\kern-0.15em\right] = 1$ if the statement is true and $\left[\kern-0.15em\left[\cdot\right]\kern-0.15em\right]=0$ otherwise. ${\rm {\bf V}}=\left( {V_{iu,jv} } \right)_{i,u,j,v}$ is a four-dimensional tensor of size $4n^2$, ${\rm {\bf c}}=\left({c_{iu} } \right)_{i,u}$ is a matrix of size $2n$. $Z$ is the normalization constant (partition function), which depends on ${\rm {\bf V}}$ and ${\rm {\bf c}}$. Note that this family has no more capacity than the original Boltzmann machine~\cite{HintonBoltzmann} and the Ising model~\cite{IsingModel}.


In Gibbs sampling without parallelism~\cite{Geman84}\cite{SABM}, we are given an observed value $y_{\mathcal{V}}$, and we want to find a sequence $\left\{y_{\mathcal{H}}^{\left( t \right)}\right\}_{t=0}^{\infty}$ which converges in distribution to the true posterior $p\left(Y_{\mathcal{H}}|Y_{\mathcal{V}}=y_{\mathcal{V}}\right)$. The algorithm proceeds as follows. First we initialize $y_{\mathcal{H}}^{\left( 0 \right)}$, and let $y_{\mathcal{V}}^{\left(0\right)}$ take on the observed values. Then at each iteration $t\in \mathbb{Z}^{+}$ we pick an index $i\in \mathcal{H}$ and do the following,
\begin{eqnarray}\label{eqn:Gibbs}
&\text{(a)}& \text{Calculate proposal } {\small \phi_{iu}^{\left( t \right)} =\sigma \left( {\sum\nolimits_{j\in M\left( i \right), v \in \left\{A, B\right\}} {W_{iu, jv} \cdot \left[\kern-0.3em\left[y_{j}^{\left( {t-1} \right)}=v\right]\kern-0.3em\right] } -b_{iu} } \right), \forall u\in \left\{A, B\right\}}\nonumber\\
&\text{(b)}& \text{Sample } y_{i}^{\left( t \right)} \sim \mbox{Bernoulli}\left( {\phi_{iA}^{\left( t \right)}, \phi_{iB}^{\left( t \right)} } \right) \text{, pass down samples by } y_{\sim i}^{\left( t \right)} =y_{\sim i}^{\left( {t-1} \right)}
\end{eqnarray}
Here we denote $W_{iu,jv} = V_{iu, jv} - V_{i\Bar{u}, jv}$ and $b_{iu} = c_{iu} - c_{i\Bar{u}}$ with $\Bar{u}$ defined as $\left\{ {\bar u} \right\} = \left\{ {A,B} \right\}\backslash \left\{ u \right\}$. $\sigma \left( x \right)=1/\left( {1+\exp \left( {-x} \right)}
\right)$ is the sigmoid function. $M\left( i \right)$ is the index set of Markov blanket of
$Y_{i} $, $i.e.$, $M\left( i \right)=\left\{ {j\in \mathcal{I}\vert j\ne i,\exists u,v \in \{ A,B\}, \text{s.t. }V_{iu,jv} \ne 0} \right\}$.

Variational inference with factorized approximation family often takes the form of a mean-field version of Gibbs sampling ({\em e.g.},~ \cite{CollapsedLDA}). We adopt the same family of variational distributions as
in \cite{DeepBM}, such that $q\left( {{Y_{\mathcal{H}}}|{\Theta _{\mathcal{H}}}} \right) = \prod\nolimits_{i \in {\mathcal{H}}} {q\left( {{Y_{i}}\vert{\theta _{iA}, \theta_{iB}}} \right)} $, where each $q\left( {Y_{i} \vert \theta_{iA}, \theta_{iB} } \right)$ is a Bernoulli distribution such that $\theta_{iu} =q\left( {Y_{i} =u\vert \theta
_{iA}, \theta
_{iB} } \right)$ for $u \in \left\{A, B\right\}$. \footnote{For simplicity, when $i\in \mathcal{V}$ we write $\theta_{iu}$ and $q\left(Y_i|\theta_{iA}, \theta_{iB}\right)$ as well, except that $\theta_{iu}$ for all iterations are fixed as the observed value. We denote ${{\mathbf \Theta }}=\left(
{\theta_{iu} } \right)_{i\in\mathcal{I}, u\in \left\{A, B\right\}}$ and write $q\left( {{\mathbf{Y}}|{\mathbf{\Theta }}} \right) = \prod\nolimits_{i \in \mathcal{I}} {q\left( {{Y_i}|{\theta _{iA}},{\theta _{iB}}} \right)} $.}
The objective of variational inference is to find $\Theta_{\mathcal{H}}$ by minimizing the loss function $L\left(
{\Theta_{\mathcal{H}}} \right)=KL\left( {q\left( {{\rm {\bf Y}}\vert {\rm
{\bf \Theta }}} \right)\vert \vert p\left( {{\rm {\bf Y}}} \right)} \right)$, here $KL\left( {q\vert \vert p} \right)=E_{q} \log \left(
{q/p} \right)$ is the KL-divergence.

The algorithm proceeds as: first we randomly initialize $\Theta_{\mathcal{H}}^{\left(0\right)}$, then at each iteration $t\in
\mathbb{Z}^{+}$ we pick an index $i\in \mathcal{H}$ and do the following,
\begin{eqnarray}\label{eqn:var}
& \text{(a)} & \text{Calculate update } \phi_{iu}^{\left( t \right)} =\sigma \left( {\sum\nolimits_{j\in M\left( i \right), v\in \left\{A, B\right\}} {W_{iu, jv} \cdot \theta_{jv}^{\left( {t-1} \right)} } -b_{iu} } \right), \forall u\in \left\{A, B\right\} \nonumber \\
& \text{(b)} & \text{Assign } \theta_{iu}^{\left( t \right)} =\phi_{iu}^{\left( t \right)}\text{, pass down }\theta_{\sim iu}^{\left( t \right)} =\theta_{\sim iu}^{\left( {t-1} \right)} , \forall u\in \left\{A, B\right\}.
\end{eqnarray}

Our motivation for combining them is as follows. First, exponential family distributions have sufficient statistics~\cite{TheoryPointEst}. By accumulating the empirical expectation of sufficient statistics, we can online estimate the parameter if data instances are completely observed, {\em e.g.}, to online estimate a Gaussian distribution $\mathcal{N}\left(X|\mu, \sigma\right)$, we can incrementally calculate $\sum\nolimits_{i = 1}^t {{x_i}}$ and $\sum\nolimits_{i = 1}^t {x_i^2}$, and estimate $\hat \mu$ and $\hat \sigma$ at each time $t$ by only using these two statistics.

Now if we take the examples in Gibbs sampling as the online data instances and estimate a distribution from within the variational approximation family, we will end up getting a marginal probability distribution for each random variable (when mixing is good). However, we can also change the way how the sequence is generated, such that the online estimated variational distribution biases towards whatever we want. In particular, we can take the calculated distribution in Eq. (\ref{eqn:var})(a) as the proposal distribution, used to sample a new data example, and incrementally update the variational distribution by using that. The updated variational distribution serve as the next input to Eq. (\ref{eqn:var}). If decaying as in~\cite{AViewEM} is used when accumulating sufficient statistics, the variational distribution estimated in our Bernoulli case, is simply a weighted sum of the most recent examples:
\begin{equation}\label{eqn:acc_theta}
\theta _{iu}^{\left( t \right)} = \sum\nolimits_{k = 0}^{ - \infty } {\epsilon \left( k \right) \cdot \left[\kern-0.3em\left[ {y_i^{\left( {t - k} \right)} = u} \right]\kern-0.3em\right]} ,{\text{  for }}u \in \left\{ {A,B} \right\}
\end{equation}
The $-\infty$ in Eq.~(\ref{eqn:acc_theta}) simply means the starting time of accumulation. $\epsilon$ function is non-negative and should sum up to one for all $k$ in the effective time span. In case constant decaying ratio is used on the last accumulated sufficient statistics as in~\cite{AViewEM}, the $\epsilon$ function above decays exponentially.
Some choices of $\epsilon$ function may not have any online updating scheme that produces it. In practice, we can either do online updating or let $\epsilon$ have a finite time course, up to some $K \in \mathbb{Z}^{+}$.


In \emph{semi-stochastic inference}, we choose a non-negative weight function ${\rm
{\epsilon}} \left(k\right)$ which sum up to 1 for $k\in \left\{1, 2, ..., K\right\}$. We
initialize both ${\rm {\bf y}}^{\left( 0 \right)}$ and ${\rm {\bf \Theta
}}^{\left( 0 \right)}$. In each iteration $t\in \mathbb{Z}^{+}$ we pick an
index $i\in \mathcal{H}$ and do the following,
\begin{eqnarray}\label{eqn:ssi_main}
& \text{(a)} & \phi _{iu}^{\left( t \right)} = \sigma \left( {\sum\nolimits_{j \in M\left( i \right),v \in \left\{ {A,B} \right\}} {{W_{iu,jv}} \cdot \theta _{jv}^{\left( t-1 \right)}}  - {b_{iu}}} \right), \text{ for } u \in \left\{ {A,B} \right\}\nonumber \\
& \text{(b)} & y_{i}^{\left( t \right)} \sim \mbox{Bernoulli}\left( {\phi_{iA}^{\left( t \right)}, \phi_{iB}^{\left( t \right)} } \right) \\
& \text{(c)} & \theta _{iu}^{\left( t \right)} = \sum\nolimits_{k = 1}^K {\epsilon \left( k \right) \cdot \left[\kern-0.3em\left[ {y_i^{\left( {t - k + 1} \right)} = u} \right]\kern-0.3em\right]} , \text{ for }u \in \left\{ {A,B} \right\} \nonumber
\end{eqnarray}
In sequential inference, for every other random variable which does not carry out these updating steps, we will pass down $\theta$ and most recent $K$ copies of $y$'s to the next time step. In step (c), additional normalization is needed when $t<K$. Details are ignored for clarity of presentation.

In another way, we can also look at algorithm in Eq.~(\ref{eqn:ssi_main}) as a ``random slowing-down'' / stochastic approximation / momentum version of variational inference. When an updated variational distribution is computed, we don't immediately turn to the updated distribution. Instead, we use it to sample a new example, and use it to incrementally update the original variational distribution. The expectation of the new example will be the same as the updated variational distribution.

\section{Making Detailed Correspondence}
\label{sec:connect}

To connect LNP neuron model in Eq.~(\ref{eqn:neuron}) with semi-stochastic inference algorithm in Eq.~(\ref{eqn:ssi_main}), several issues needs to be resolved:
\begin{enumerate}
\item \label{iss:nobias} LNP neurons don't have the bias terms (such as $b_{iu}$), although their nonlinear transform function may have constant shift identical to all neurons.
\item \label{iss:event} In Eq.~(\ref{eqn:ssi_main}), two Linear-Nonlinear procedures in (a)(c) corresponds to only one sampling step in (b). This is an architectural difference.
\item \label{iss:eionly} Biological neurons can only be excitatory or inhibitory.
\item \label{iss:par} Biological neurons carry out the updates all in parallel.
\end{enumerate}
To resolve issue~\ref{iss:nobias}, we notice that when the weights $\mathbf{V}$ and biases $\mathbf{c}$ in Eq.~(\ref{eq4}) are turned into $\mathbf{W}$ and $\mathbf{b}$ in Eq.~(\ref{eqn:ssi_main}), for fixed $i,j \in \mathcal{I}$, fixed $u,v \in \left\{A, B\right\}$, and any constant $C \in \mathbb{R}$ the following operation is invariant to the inference algorithm in Eq.~(\ref{eqn:ssi_main}) since $\theta_{iA}^{\left(t\right)}$ and $\theta_{iB}^{\left(t\right)}$ sum up to one.
\begin{equation}
\left\{ \begin{gathered}
  b_{iu}^{\text{new}} \leftarrow b_{iu}^{\text{old}} + C \hfill \\
  W_{iu,jv}^{\text{new}} \leftarrow W_{iu,jv}^{\text{old}} + C \hfill \\
  W_{iu,j\bar v}^{\text{new}} \leftarrow W_{iu,j\bar v}^{\text{old}} + C \hfill \\
\end{gathered}  \right.
\end{equation}
For this reason, we can always modify the algorithm in Eq.~(\ref{eqn:ssi_main}) to make any bias we want. In our experiments, when we want to remove a bias $b_{iu}$, we subtract $b_{iu}/ {\vert M\left(i\right)\vert}$ from both $W_{iu,jv}$ and $W_{iu,j\Bar{v}}$ for each $j \in M\left(i\right)$.

To resolve issue~\ref{iss:event}, we propose the ``Event-Network''. We first split every sampling step (b) in Eq.~(\ref{eqn:ssi_main}) into two, namely $[\kern-0.15em[ {y_{i}^{\left( t \right)} =A} ]\kern-0.15em]\sim \mbox{Bernoulli}( {\phi_{iA}^{\left( t \right)} } )$ and $[\kern-0.15em[ {y_{i}^{\left( t \right)} =B} ]\kern-0.15em]\sim \mbox{Bernoulli}( {\phi_{iB}^{\left( t \right)} } )$. Now we can allocate two neurons to take care of the two parts in all three steps (a)(b)(c). Units of the resulted network no longer corresponds to random variables, they are now probabilistic events. When we do this, there is no guarantee that $\theta_{iA}^{\left(t\right)}$ and $\theta_{iB}^{\left(t\right)}$ will sum up to 1, but this will be approximately true since the weights $W_{iu,jv}$'s are compatible with each other. Whether this results in meaningful inference algorithms needs further investigation. We show verification in our experiment part.


To resolve issue~\ref{iss:eionly}, once we have a network containing neurons with both excitatory and inhibitory outgoing synapses, we can duplicate each neuron into two, both have the same incoming synapses with the original neural efficacy. Then one of them take all positive outgoing synapses before splitting, and the other take all negative ones. Both of this and the last modification of the network are exactly invariant for variational inference but may only be approximately true for semi-stochastic inference. We show experimental verification for this modification as well.

If after resolving all these issues we denote $x_{iu}^{\left(t\right)} = [\kern-0.15em[ {y_i^{\left( t \right)} = u} ]\kern-0.15em]$ for $u\in \left\{A, B\right\}$, and re-index everything, the resulted algorithm will be exactly the same as Eq.~(\ref{eqn:neuron}) if executed in parallel. Parallelism is not only an issue in semi-stochastic inference. It is noticed in~\cite{SABM} that when adjacent random variables of a Boltzmann machine are updated in parallel in Gibbs sampling, the sample sequence does not converge to the right distribution. This issue is also raised very often recently ({\em e.g.} \cite{ParGibbs}). Variational inference is a fixed point iteration. So it may converge to the same answer when updated all in parallel, but we do observed experimentally that the variational parameters converges to the fixed point after severe oscillation. Whether the slowing down version alleviates this problem requires further investigation\footnote{We observe that different choices of $\epsilon$ functions result in different smoothness of the trajectories, but it's not clear how. We leave this as future work.}.

\section{Experimental Verification}
\label{sec:inf_dbm}

\begin{figure}[h]
\includegraphics[width=1\linewidth]{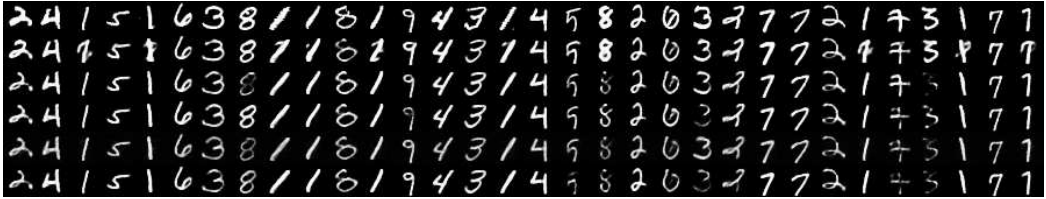}
\caption{Samples of reconstruction results (selected randomly). Rows from top to bottom: original images; \textbf{VarO}; \textbf{SemiO}; \textbf{SemiEN}; \textbf{SemiB}; \textbf{SemiU}.}
\label{fig:rec}
\end{figure}
In this section, we focus on illustrating that the semi-stochastic inference, as well as all the modifications in section~\ref{sec:connect}, are valid Bayesian inference algorithms, which behave similar to variational inference. In all experiments, we set $\epsilon(k)$ as $\exp(-k/2)$ normalized to have summation 1 over its domain $k\in \left\{1, ..., 30\right\}$. All inferences are carried out in synchronized full parallelism.

To avoid the complication of learning deep Boltzmann machines, we take the learned Boltzmann machine from code in~\cite{DeepBM}. Since we focus on inference on Boltzmann machine, we take the learned model before back-propagation. The Boltzmann machine we use consists of three layers, the first layer for input image consists of 784 units, the second consists of 500 units and the third consists of 1000 units. In the following, we refer to $\theta_i^{(t)}$ in both algorithms as the activation of neuron $i$.

To see the effect of modifications in section~\ref{sec:connect}, we show results of semi-stochastic inference after each of them. Note that variational inference is exactly invariant on all these modifications, hence there no need to verify them. We use the following abbreviations in this part:
\begin{itemize}
\item \textbf{VarO:} variational inference applied on the original deep Boltzmann machine.
\item \textbf{SemiO:} semi-stochastic inference applied on the original deep Boltzmann machine.
\item \textbf{SemiEN:} Same as \textbf{SemiO} except each sampling is duplicated and put on two neurons.
\item \textbf{SemiB:} Same as \textbf{SemiEN} except biases are removed by transforming the network.
\item \textbf{SemiU:} Same as \textbf{SemiB} except each neuron is duplicated into two, such that no neuron contains out-going synapses with different signs.
\end{itemize}

The experiment we did is similar to the reconstruction experiment in~\cite{Hinton06DimRed}. We do the following: (i) plug in an image on the input layer, (ii) apply the inference algorithm, (iii) frozen the activations on the topmost layer and round up to 0 or 1, (iv) set the topmost layer as observed and other layers as hidden, (v) infer back, (vi) read off the image as activations on the input layer. Both (ii) and (v) are randomly initialized and have 100 iterations. For semi-stochastic inference, in steps (iii) and (vi) the activations are taken as averages over the last 50 iterations, while for variational inference we simply use the last activation of each neuron. This is because semi-stochastic inference has random convergence, activations approach some fixed point with random perturbation (see justification later). Some reconstruction examples are shown in Fig.~\ref{fig:rec}. Each algorithm has good and bad cases, also note that because of the randomness, even for identical input image and identical initialization, semi-stochastic inference may get different results in different trials.

\begin{figure}[t]
\begin{center}
\begin{tabular}{cccc}
\includegraphics[width=0.22\linewidth]{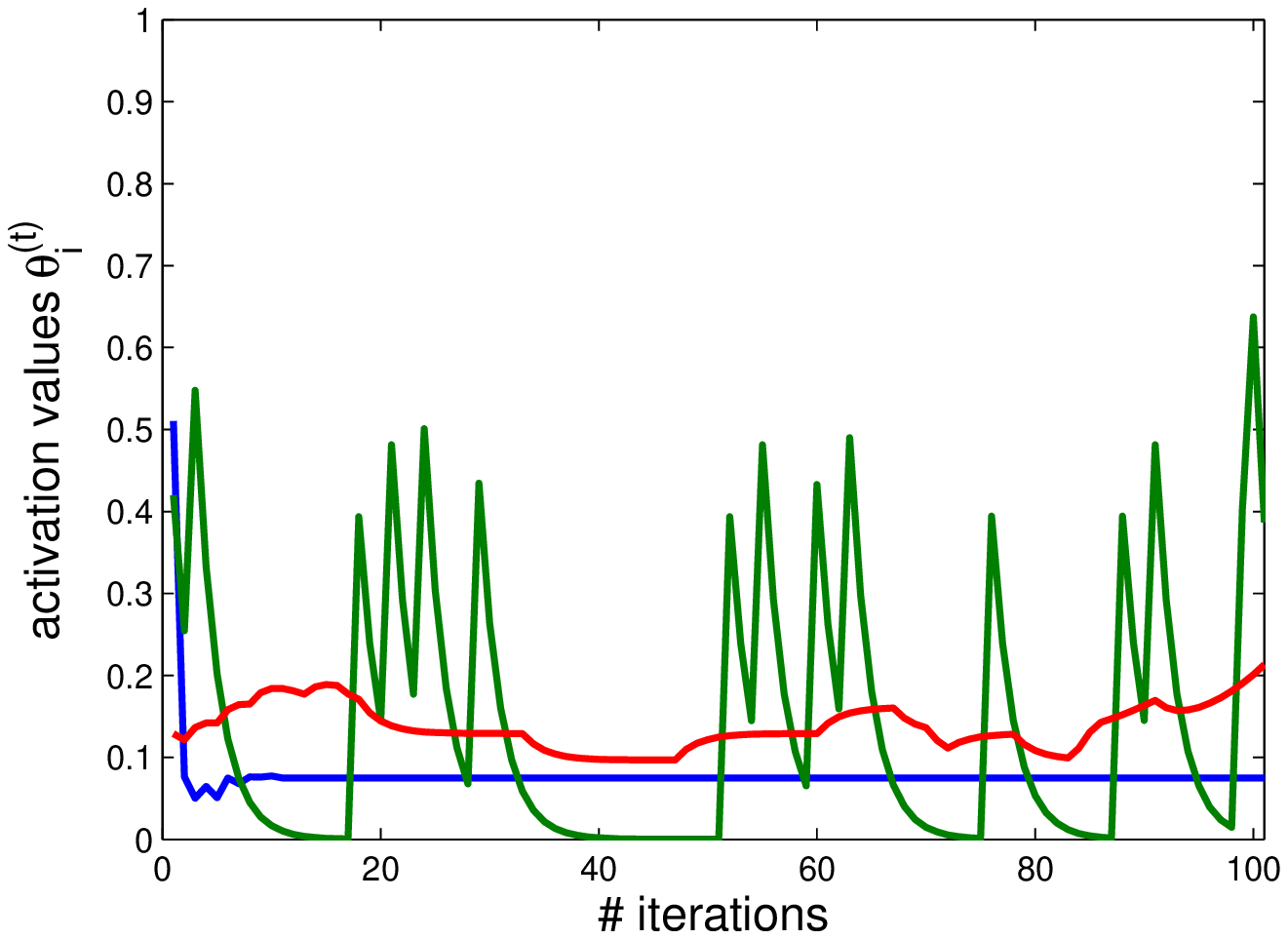} &
\includegraphics[width=0.22\linewidth]{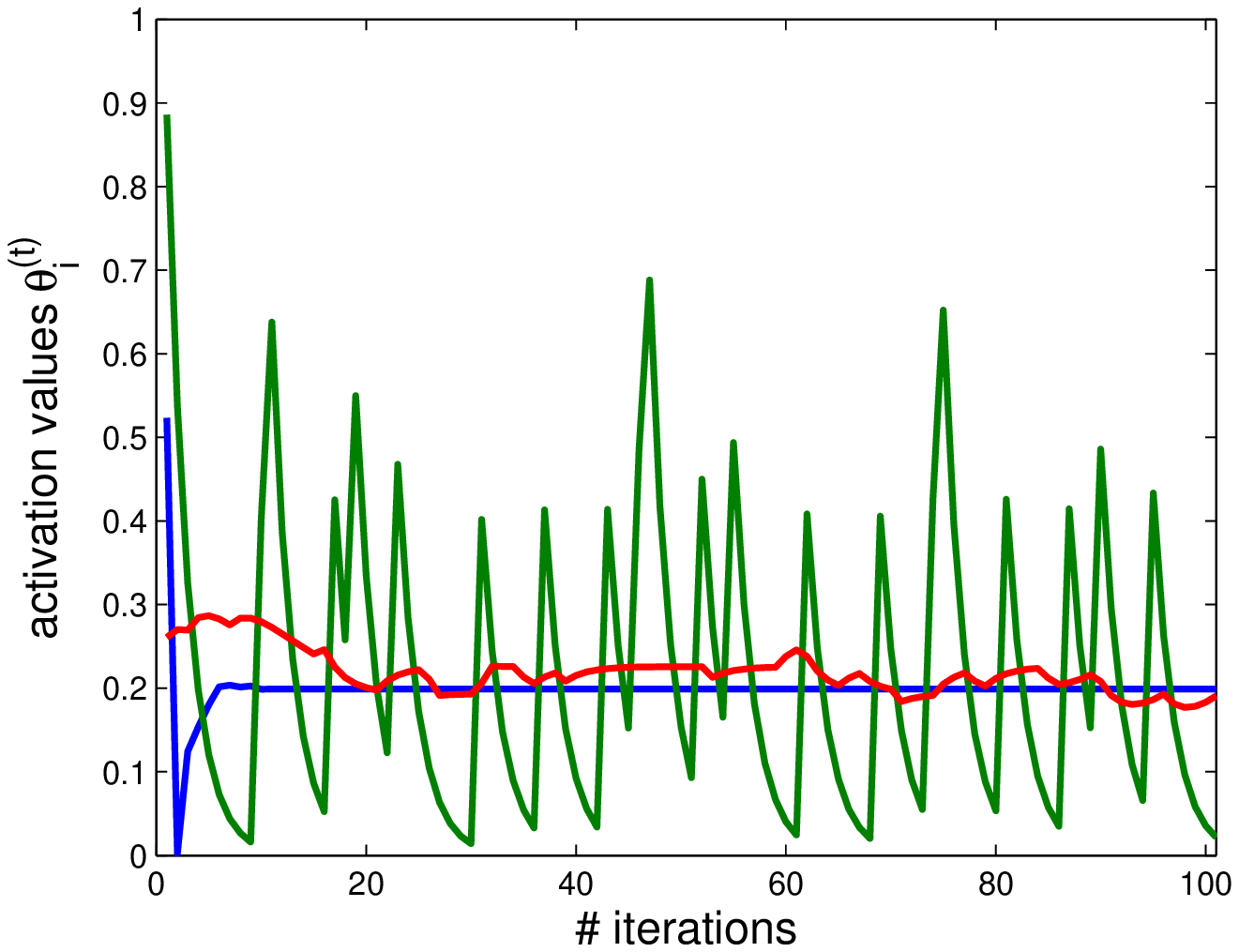} &
\includegraphics[width=0.22\linewidth]{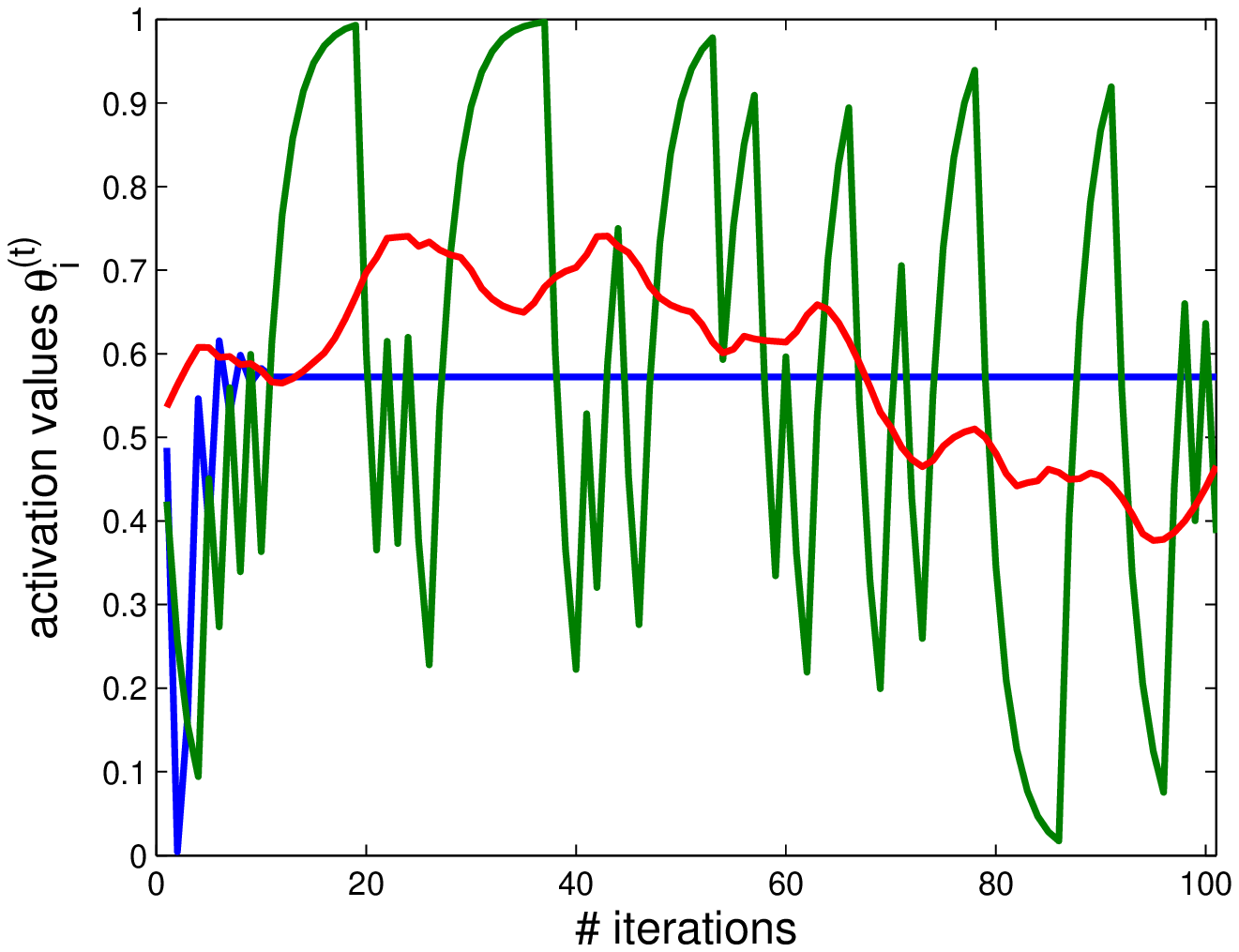} &
\includegraphics[width=0.22\linewidth]{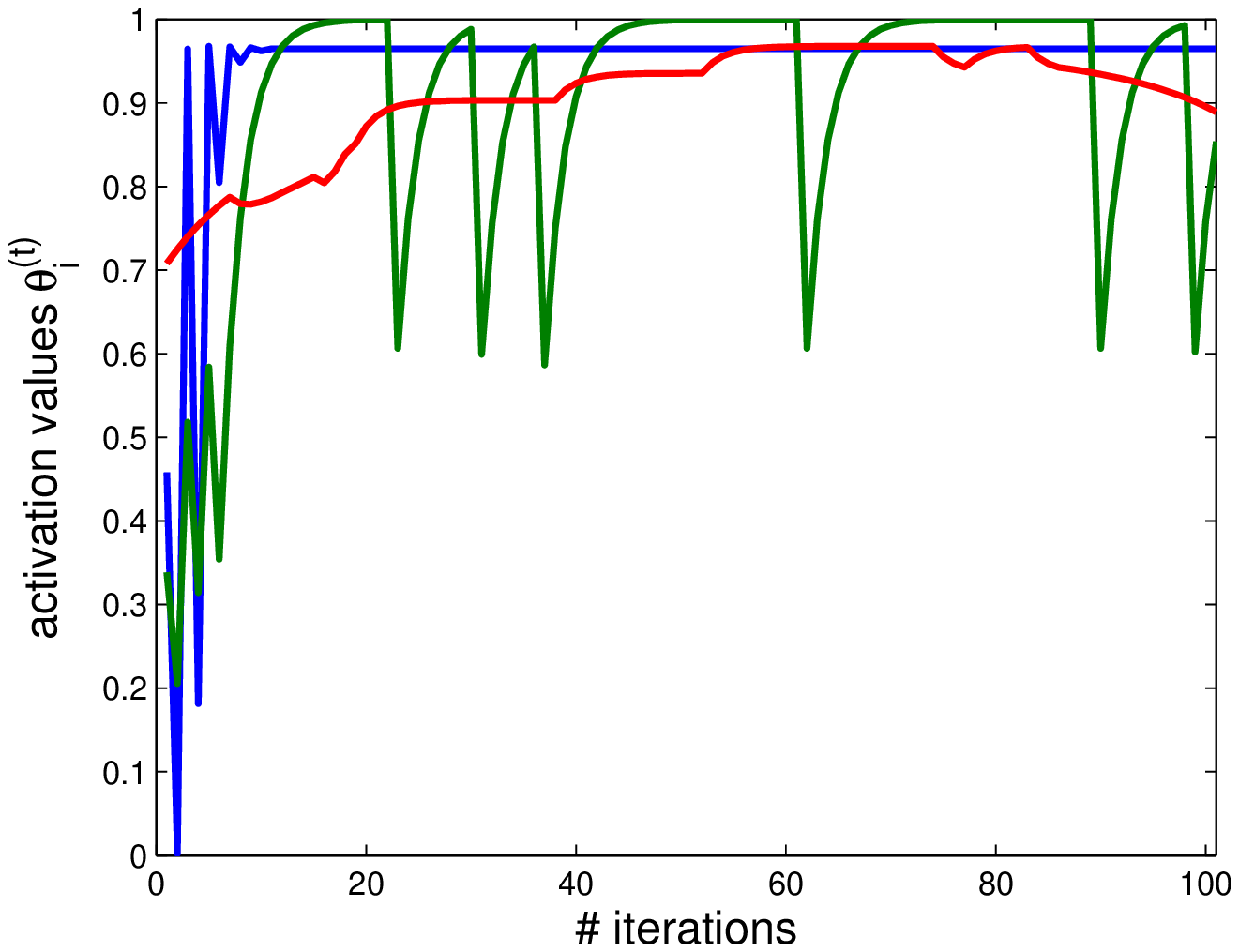}
\end{tabular}
\end{center}
\caption{Sample trajectory of $\theta_i^{(t)}$ in \textbf{VarO} and \textbf{SemiO}. In all above, \textcolor[rgb]{0.00,0.00,1.00}{Blue} line is for \textbf{VarO}, \textcolor[rgb]{0.00,0.50,0.00}{Green} line is for \textbf{SemiO}, and \textcolor[rgb]{1.00,0.00,0.00}{Red} line is the moving average of Green line over window size 30.}
\label{fig:varssi}
\end{figure}
In figure~\ref{fig:varssi}, we showed sample trajectory of activations in step (ii) of \textbf{VarO} and \textbf{SemiO} for the same variables. Each trajectory are those $\theta_i^{(t)}$'s of one neuron over all iterations. We see that activations in variational inference converge to a stable solution with oscillation before convergence, while semi-stochastic inference has their moving average converges to some similar values with random perturbation. Figure~\ref{fig:stat}(a) is a scatter plot of mean of trajectories vs. the corresponding converged activation in variational inference, extracted from 2000 trajectories randomly chosen from 50 trials. We also observed that when the mean is close to extreme values (such as 0 and 1), the standard variance will be smaller, as can be found in figure~\ref{fig:varssi} and more clearly in figure~\ref{fig:stat}(b). This means that semi-stochastic inference converges randomly to the variational inference solution, with more confident variables having less random perturbation.

\begin{figure}[t]
\begin{center}
\begin{tabular}{ccc}
\includegraphics[width=0.30\linewidth]{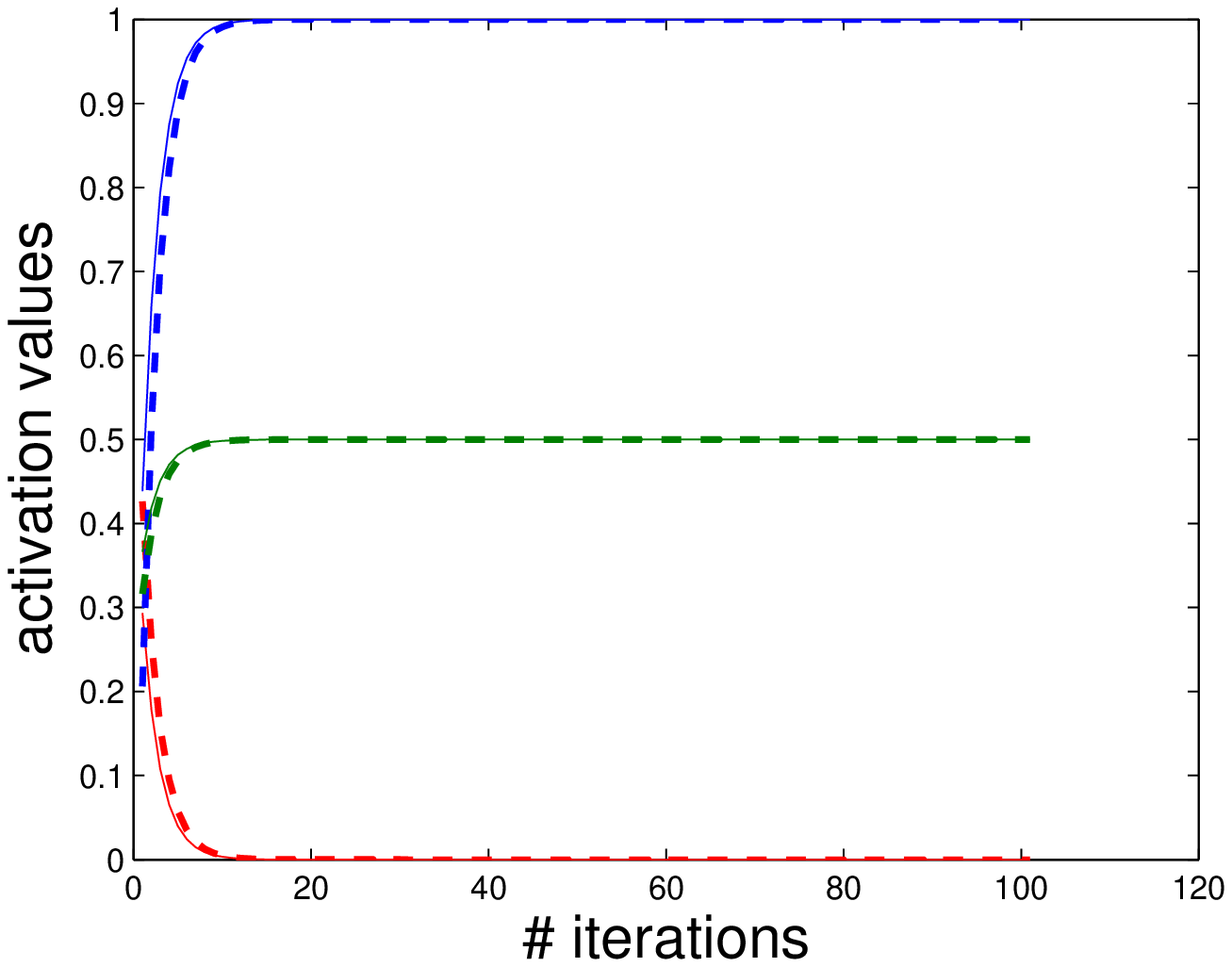} &
\includegraphics[width=0.30\linewidth]{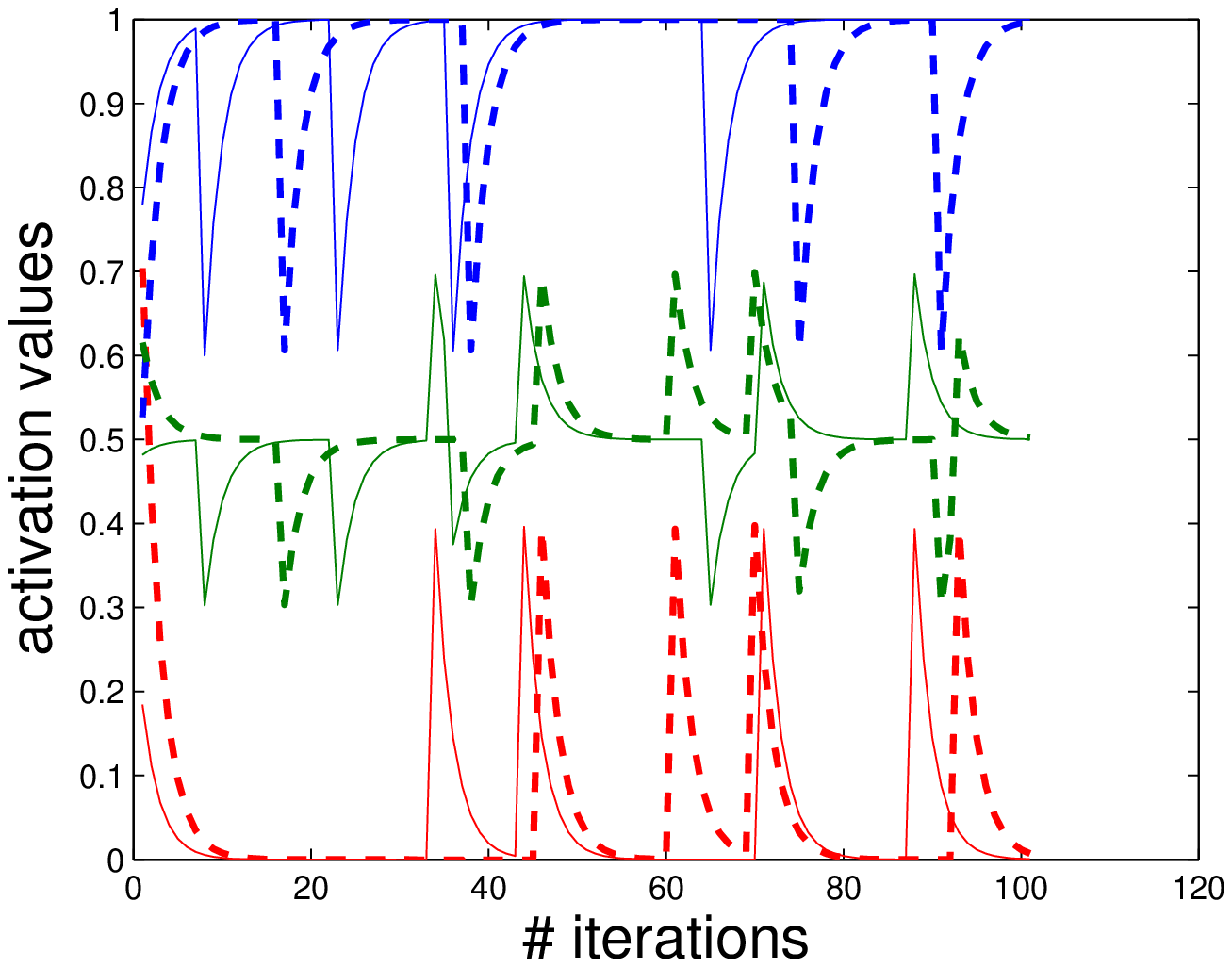} &
\includegraphics[width=0.30\linewidth]{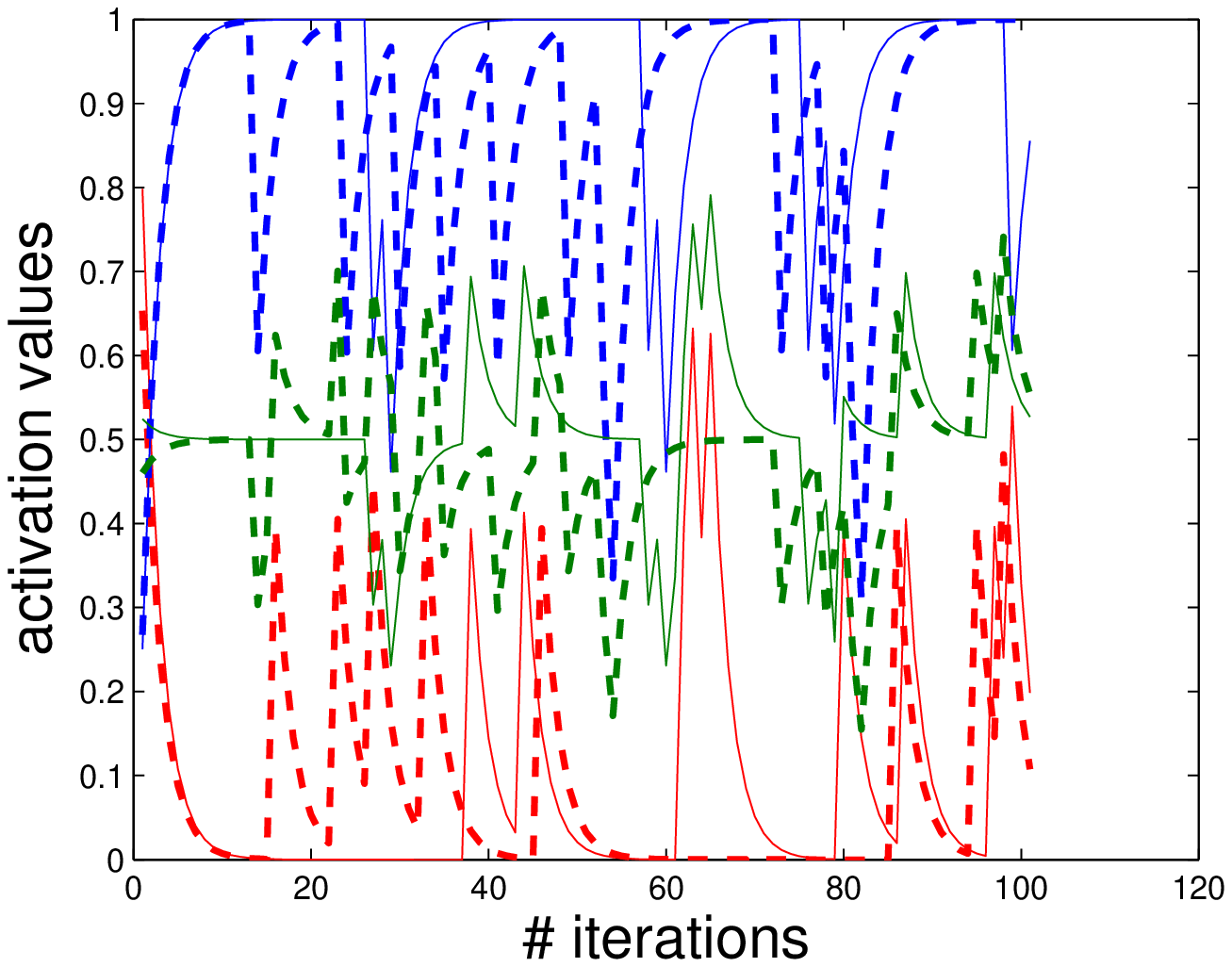} \\
(a) & (b) & (c)
\end{tabular}
\end{center}
\caption{Sample trajectory from \textbf{SemiU}. Each plot shows four neurons resulted from two splits on a single neuron. Thick dash lines: neurons with positive outgoing weights. Thin solid lines: corresponding neurons with negative outgoing weights. \textcolor[rgb]{0.00,0.00,1.00}{Blue} and \textcolor[rgb]{1.00,0.00,0.00}{Red} lines are for the two neurons from splitting one sampling into two. \textcolor[rgb]{0.00,0.50,0.00}{Green} line is the average of Blue and Red. Ideally, lines with same color should be equal, and green lines should be at 0.5.}
\label{fig:semiu}
\end{figure}

\begin{figure}[t]
\begin{center}
\begin{tabular}{cccc}
\includegraphics[width=0.22\linewidth]{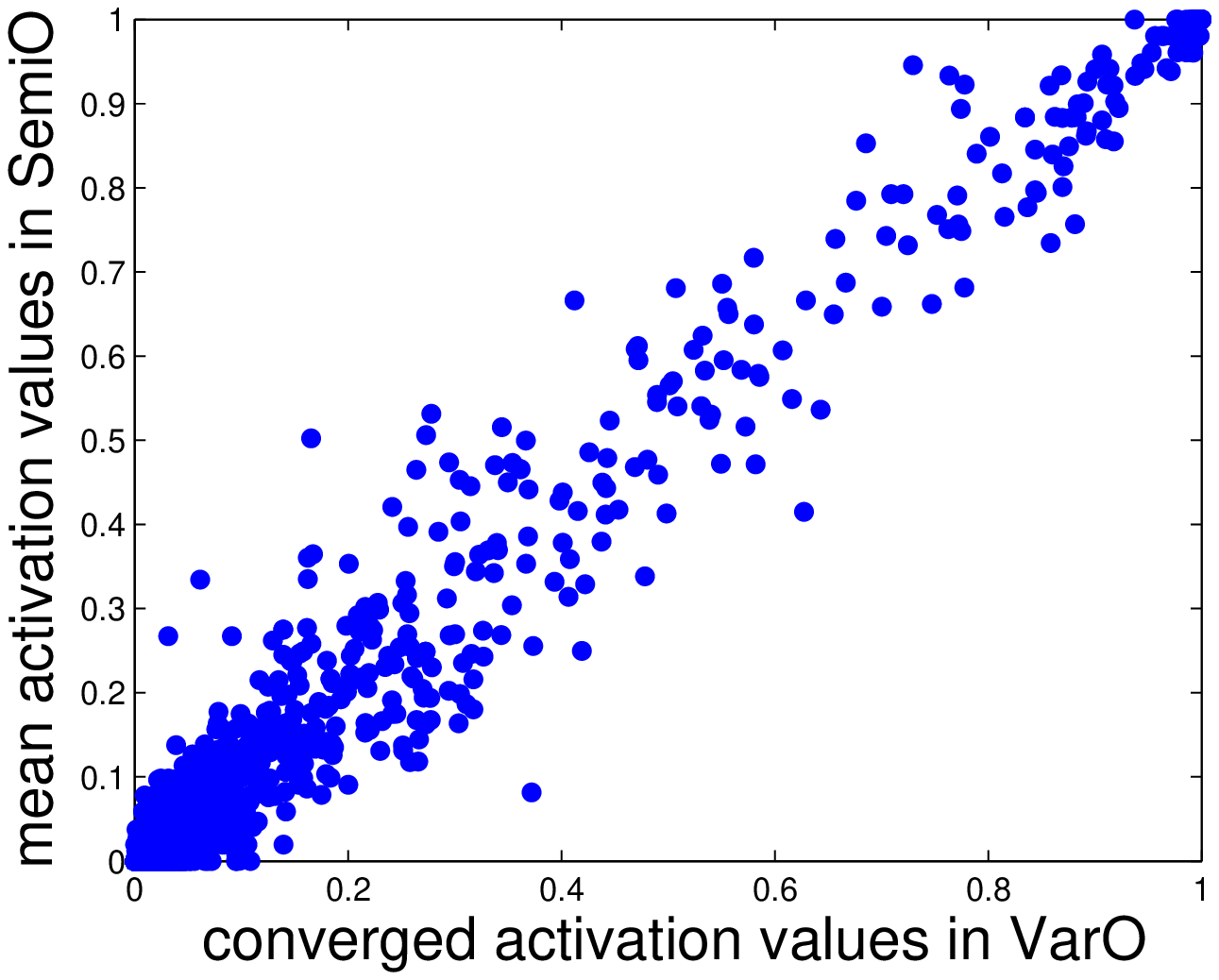} &
\includegraphics[width=0.22\linewidth]{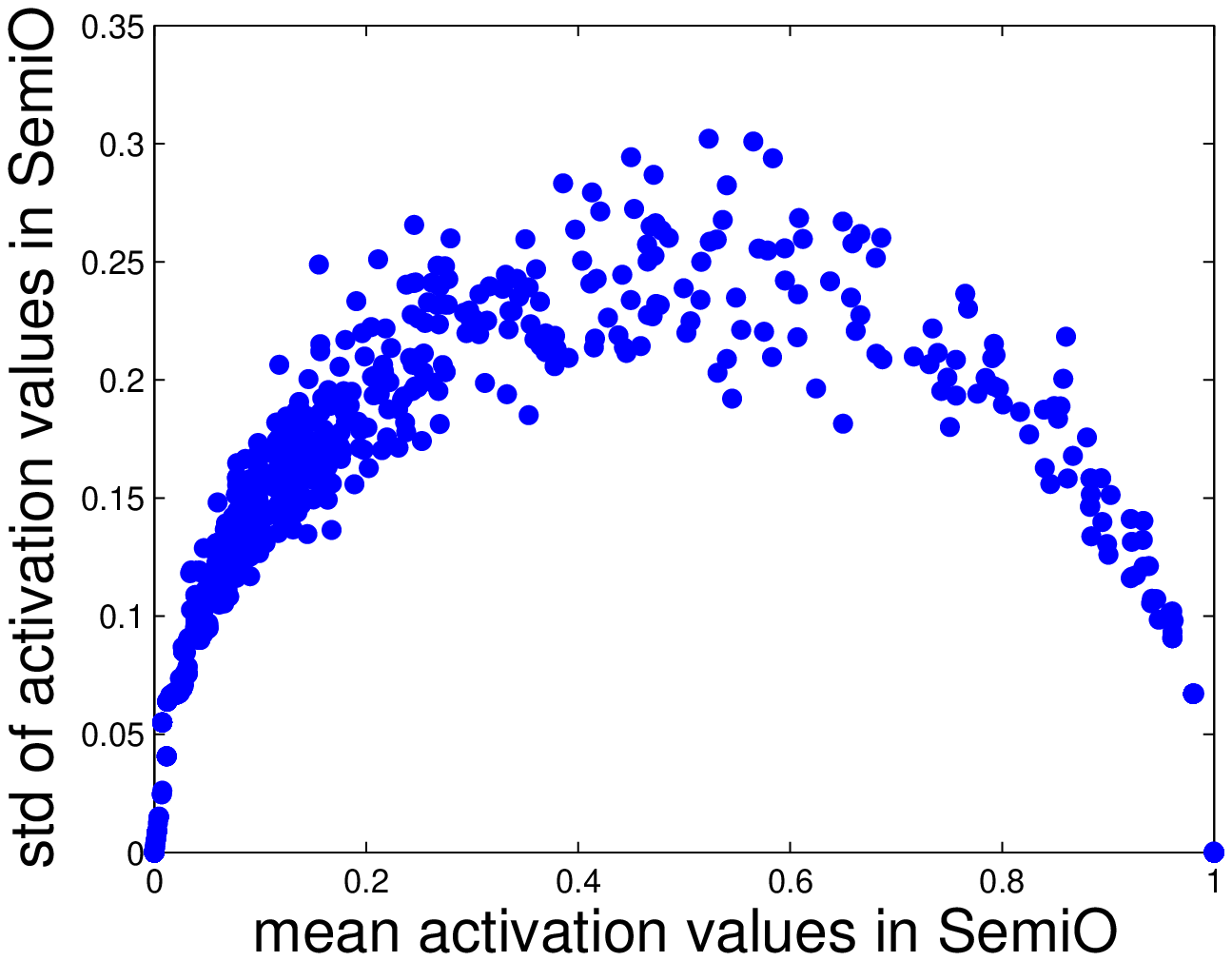} &
\includegraphics[width=0.22\linewidth]{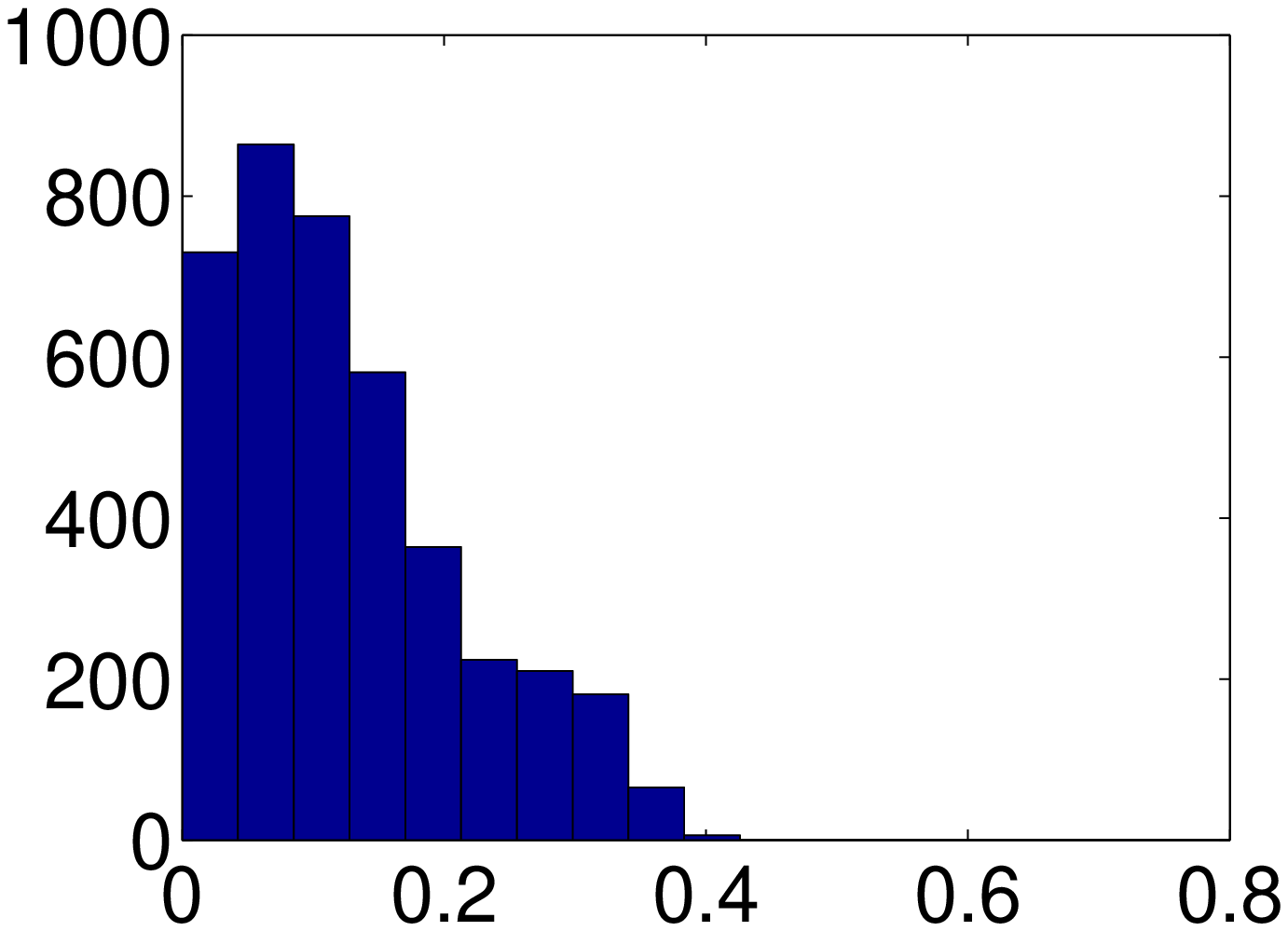} &
\includegraphics[width=0.22\linewidth]{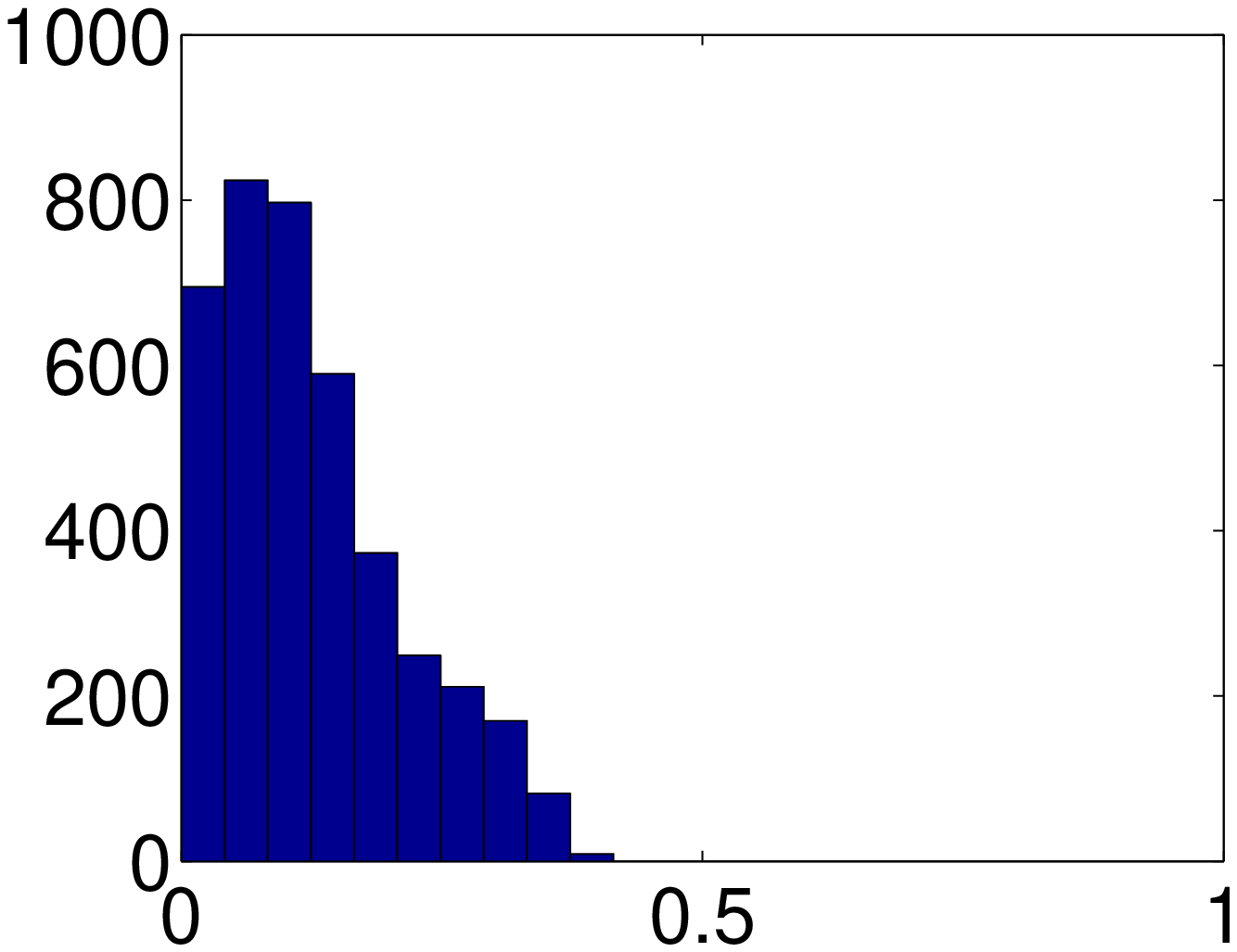} \\
(a) & (b) & (c) & (d)
\end{tabular}
\end{center}
\caption{Some statistics of the algorithms. (a) means of trajectories in \textbf{SemiO} vs. converged activation values in \textbf{VarO} for the same hidden variables. (b) standard deviation vs. mean, from the trajectories in \textbf{SemiO}. (c) histogram of standard\_deviation(red line + blue line - 1.0). (d) histogram of standard\_deviation(dash line - solid line). (Here red, blue, dash, solid lines refer to figure~\ref{fig:semiu}.)}
\label{fig:stat}
\end{figure}

Finally, we investigated on how well the splitting in section~\ref{sec:connect} preserves identity relationships that are expected on \textbf{SemiU}. In particular, if two neurons are resulted from splitting one sampling into two in resolving issue~\ref{iss:event}, their summed activation should be close to 1. If two neurons are resulted from splitting one neuron into two in resolving issue~\ref{iss:eionly}, their activation should be equal. Figure~\ref{fig:semiu} illustrated three samples and histogram on 2000 random trajectories are given in figure~\ref{fig:stat}(c)(d).


\section{Conclusions and Discussions}


We pointed out the stochastic inference nature of the LNP neuron models, which may serve as an interpretation of neural coding. The stochastic convergence seems reasonable, {\em e.g.}, when we see ambiguous images, our perception will switch back and forth between plausible explanations.

There are many behavioral experiments showing statistical optimality of perception and learning ({\em e.g.} \cite{ProbNeuron}) and many calls for neuronal modeling that achieves this optimality ({\em e.g.} \cite{FiserVisual}). Arguably, the mode-seeking nature of variational inference~\cite{DivergenceMeasure} make it natural for interpreting perception. Since Boltzmann machines with hidden variables are universal approximators ({\em e.g.} \cite{BengioRepresentationalPower}), modeling knowledge representation by that is safe if the world can be approximately binarized. Yet we only show that with particular choices of weights, neurons are capable of representing a Boltzmann machine and carry out inference. Then the question is what does this do if the network is a recurrent neural network in general. Real biological neural networks may be learned by taking the Boltzmann machine representation and undergoing discriminative training or reinforcement learning to optimize performance of inference directly (especially plausible when time dimension is involved), which yields a model not necessarily convey consistent probabilistic semantics but may do so in the limit. The particular form of nonlinearity in biological neurons may also result from such optimization. Another issue is the low average activity in biological neurons. It may be a certain form of sparse coding~\cite{OlshausenSparseCoding} on top of inference so that optimizing such inference procedure will not yield degenerate solution. In this case, likelihood-based learning such as contrastive divergence may still be applicable as a further refinement.

\section*{Acknowledgement}

I want to thank Geoffrey Hinton, Mikhail Belkin, DeLiang Wang, Brian Kulis for helpful discussion and The Ohio Supercomputer Center for providing computation resource.

{
\bibliography{BibTex_YuanlongShao}
\bibliographystyle{unsrt}
}

\end{document}


\maketitle


In this supplemental material, we deal with two issues that are not made
clear in the main text: (\ref{eq1}) we give the details of discrete time
approximation, which eventually leads to the inference algorithm; (\ref{eq2}) we
present preliminary results on stochastic stability of the neuron model.
After that, we also point out several open questions related to the
Linear-Nonlinear-Poisson (LNP) neuron model and its interpretation of
Bayesian inference, with possible directions for solving these problems.

For now, we focus on the neuron model with a Poisson Point Process (PPP)
when there is no linear convolution with function $D$ in the main text and
the dendritic integration is linear, as the following
\begin{equation}
\label{eq1}
\begin{array}{l}
 \lambda_{i} \left( \tau \right)=\sigma \left( {\sum\nolimits_{j\in M\left(
i \right)} {W_{ij} \cdot \left( {X_{j} \ast \alpha } \right)\left( \tau
\right)} +e_{i} } \right), \\
 X_{i} \left( \tau \right)\sim \text{PoissonPointProcess}\left( {\lambda
_{i} \left( \tau \right)} \right),\text{\thinspace \thinspace \thinspace
\thinspace \thinspace \thinspace }\forall i\in \mathcal{I} \\
 \end{array}
\end{equation}
Here $\tau \in \mathbb{R}$ is the continuous time index, which has unit
millisecond. An external input term $e_{i} $ is introduced so that we don't
need to explicitly distinguish visible and hidden units. Every $e_{i} $ is
constant over time. For example, $e_{i} =0$ means the neuron corresponds to
a hidden unit, while a very positive and a very negative $e_{i} $ correspond
to a visible unit with observation 1 and 0 respectively. We also assume that
$\tau =0$ is the beginning time and each $\lambda_{i} \left( 0 \right)$ is
chosen by the user

\section{The Stochastic Differential Equation Formulation}
\label{sec:mylabel1}
In this section, we reformulate Eq. (\ref{eq1}) as a stochastic differential
equation, which is for the ease of the later parts.

If we choose the parametrization of $\alpha $ function as the exponential
function $\alpha \left( \tau \right)=a\cdot \exp \left( {-a\cdot \tau }
\right)$, and define $Y_{i} \left( \tau \right)\triangleq \left( {X_{i} \ast
\alpha } \right)\left( \tau \right)$, then $Y_{i} $ is the solution of the
differential equation as the following.
\begin{equation}
\label{eq2}
\frac{dY_{i} }{d\tau }=-a\cdot Y_{i} +a\cdot X_{i}
\end{equation}
Thus we can rewrite Eq. (\ref{eq1}) in terms of ${\mathbf{Y}}=\left( {Y_{i} }
\right)_{i\in \mathcal{I}} $, and we get a system of coupled stochastic
differential equations (SDE) as the following
\begin{equation}
\label{eq3}
\begin{array}{l}
 \frac{d{\mathbf{Y}}}{d\tau }=-a\cdot \left( {{\mathbf{Y}}-{\mathbf{\lambda
}}} \right)+a\cdot \left( {{\mathbf{X}}-{\mathbf{\lambda }}}
\right),\text{\thinspace \thinspace where} \\
 \forall i\in \mathcal{I},\text{\thinspace \thinspace }\left\{
{\begin{array}{l}
 \lambda_{i} \left( \tau \right)=\sigma \left( {\sum\nolimits_{j\in M\left(
i \right)} {W_{ij} \cdot Y_{j} \left( \tau \right)} +e_{i} } \right) \\
 X_{i} \left( \tau \right)\sim \text{PoissonPointProcess}\left( {\lambda
_{i} } \right) \\
 \end{array}} \right. \\
 \end{array}
\end{equation}
The right hand side of Eq. (\ref{eq3}) contains two terms, the deterministic term
$-a\cdot \left( {{\mathbf{Y}}-{\mathbf{\lambda }}} \right)$ and the
stochastic term $a\cdot \left( {{\mathbf{X}}-{\mathbf{\lambda }}} \right)$.
The deterministic term reaches zero when ${\mathbf{Y}}={\mathbf{\lambda }}$,
and thus is the equilibrium of the (recurrent) neural network, if there is
one. The stochastic part, roughly speaking, has short time integral close to
zero, which is referred to as \textit{random perturbation}. The
deterministic counterpart of Eq. (\ref{eq3}) is defined as the following.
\begin{equation}
\label{eq4}
\frac{d{\mathbf{Y}}}{d\tau }=-a\cdot \left( {{\mathbf{Y}}-{\mathbf{\lambda
}}} \right)
\end{equation}
In the literature of stochastic differential equations, the stochastic
stability of Eq. (\ref{eq3}) is defined in many sense analogous to the Lyapunov
stability of Eq. (\ref{eq4}). For example, in \cite{StabSDE} the \textit{stability
under small random perturbation} is introduced, which is defined as a
capability of the deterministic system, in the sense that as long as the
random force can be made sufficiently small, the combined stochastic system
will exhibit certain stability, although the requirement imposed upon the
strength of the random force may not satisfy It is likely that Eq. (\ref{eq4})
satisfies this stability property. We haven't obtained a rigorous proof yet,
but in the following part, we give an analysis of this kind of stochastic
stability in the discrete time case. We first give the detail about how the
discrete case is achieved and then proceed by analyzing it.

\section{Details of Discrete Time Approximation}
\label{sec:details}
In this section, we give the details about how the PPP formulation is
discretized into the Bernoulli process formulation, and how it is related to
the fixed-point iteration procedure, which are not made clear in the main
text.

Let $t\in \mathbb{Z}$ be the discrete time index such that discrete time $t$
corresponds to continuous time $\tau =t\epsilon $, where $\epsilon \in
\mathbb{R}^{+}$ is the length of a small time step. The discrete time
approximation of PPP in Eq. (\ref{eq3}) is as the following.
\begin{equation}
\label{eq5}
x_{i}^{\left( t \right)} \sim \text{Bernoulli}\left( {\epsilon \cdot \lambda
_{i}^{\left( t \right)} } \right),\text{\thinspace \thinspace
where\thinspace }\lambda_{i}^{\left( t \right)} =\frac{1}{\epsilon
}\int_{\left[ {t\epsilon ,t\epsilon +\epsilon } \right)} {\lambda_{i}
\left( \tau \right)d\tau }
\end{equation}
By LeCam's theorem \cite{LecamPoisson}, it can be shown that when $\epsilon
\to 0$, for any fixed time interval $\left[ {\tau_{1} ,\tau_{2} }
\right)$, the sum of $x_{i}^{\left( t \right)} $ fall in that interval
converges in distribution to a Poisson distribution with rate $\int_{\tau
_{1} }^{\tau_{2} } {\lambda_{i} \left( \tau \right)d\tau } $. That means
the non-homogeneous Bernoulli process approaches non-homogeneous PPP We
define $y_{i}^{\left( t \right)} $ as the following (in which we ignored the
notation issue regarding the starting point)
\begin{equation}
\label{eq6}
y_{i}^{\left( t \right)} =\sum\nolimits_{k=0}^\infty {\alpha^{\left( k
\right)}x_{i}^{\left( {t-1-k} \right)} } ,\text{\thinspace \thinspace
where\thinspace }\alpha^{\left( k \right)}=a\cdot \left( {1-a\epsilon }
\right)^{k}
\end{equation}
With this definition, we have the following approximation formula
\begin{equation}
\label{eq7}
\alpha^{\left( {\tau /\epsilon } \right)}=a\cdot \left( {1-a\epsilon }
\right)^{\tau /\epsilon }\to a\cdot \exp \left( {-a\tau } \right)=\alpha
\left( \tau \right),\text{\thinspace \thinspace as\thinspace }\epsilon \to
0
\end{equation}
For this reason, $y_{i}^{\left( t \right)} $ defined in Eq. (\ref{eq6}) is an
approximation of $Y_{i} \left( {t\epsilon } \right)$ when the Bernoulli
process is an approximation of PPP. Eq. (\ref{eq6}) can also be rewritten
recursively as the following.
\begin{equation}
\label{eq8}
y_{i}^{\left( {t+1} \right)} =\left( {1-a\epsilon } \right)\cdot
y_{i}^{\left( t \right)} +a\cdot x_{i}^{\left( t \right)}
\end{equation}
In addition, for $\tau >0$, each $Y_{i} \left( \tau \right)$ is right
continuous with discontinuity located only at the spiking points indicated
by $X_{i} \left( \tau \right)$, and so is $\lambda_{i} \left( \tau
\right)$. Thus when $\epsilon \to 0$, $\lambda_{i}^{\left( {\left\lceil
{\tau /\epsilon } \right\rceil } \right)} \to \lambda_{i} \left( \tau
\right)$, where $\left\lceil {\tau /\epsilon } \right\rceil $ is the
smallest integer that is larger than $\tau /\epsilon $. For this reason,
$\lambda_{i}^{\left( t \right)} $ is an approximation of $\lambda_{i}
\left( {t\epsilon } \right)$. Since $\lambda_{i} \left( {t\epsilon }
\right)$ is a function of $Y_{i} \left( {t\epsilon } \right)$ and
$y_{i}^{\left( t \right)} $ approximates $Y_{i} \left( {t\epsilon }
\right)$, $\lambda_{i}^{\left( t \right)} $ is approximately a function of
$y_{i}^{\left( t \right)} $.

As a conclusion, the discrete time approximation of Eq. (\ref{eq3}) is the
following, where the first line is derived from Eq. (\ref{eq8}). The approximation
gets better (closer to the continuous time counterpart) when $\epsilon \to
0$
\begin{equation}
\label{eq9}
\begin{array}{l}
 \frac{{\mathbf{y}}^{\left( {t+1} \right)}-{\mathbf{y}}^{\left( t
\right)}}{\epsilon }=-a\cdot \left( {{\mathbf{y}}^{\left( t
\right)}-{\mathbf{\lambda }}^{\left( t \right)}} \right)+a\cdot \left(
{{\mathbf{x}}^{\left( t \right)}/\epsilon -{\mathbf{\lambda }}^{\left( t
\right)}} \right),\text{\thinspace \thinspace where} \\
 \lambda_{i}^{\left( t \right)} =\sigma \left( {\sum\nolimits_{j\in M\left(
i \right)} {W_{ij} \cdot y_{j}^{\left( t \right)} } +e_{i}^{\left( t
\right)} } \right),\text{\thinspace \thinspace }x_{i}^{\left( t \right)}
\sim \text{Bernoulli}\left( {\epsilon \cdot \lambda_{i}^{\left( t \right)}
} \right) \\
 \end{array}
\end{equation}
We can also replace the first formula in Eq. (\ref{eq9}) by Eq. (\ref{eq6}). The resulted
formula matches the neuron model in our main text. Again we can define the
deterministic counterpart of Eq. (\ref{eq9}) as the following.
\begin{equation}
\label{eq10}
\frac{{\mathbf{y}}^{\left( {t+1} \right)}-{\mathbf{y}}^{\left( t
\right)}}{\epsilon }=-a\cdot \left( {{\mathbf{y}}^{\left( t
\right)}-{\mathbf{\lambda }}^{\left( t \right)}} \right)
\end{equation}
Note that when $a\epsilon =1$, Eq. (\ref{eq10}) is exactly the forward evaluation
step of (recurrent) neural networks. When $a\epsilon <1$, it is the momentum
version of the forward step. In case the neural network represents the
variational inference algorithm over Boltzmann machines, \textit{i.e.},
weights are symmetric, the forward step is guaranteed to converge to a local
optimum, no matter where the starting point is.

\section{Stochastic Stability of the Neuron Model}
\label{sec:stochastic}
In this section, we illustrate the stochastic stability of Eq. (\ref{eq9}).
Eventually, the stability of the stochastic system will rely on the
stability of the deterministic system as in Eq. (\ref{eq10}). In general it is not
clear when the dynamical system, either discrete time in Eq. (\ref{eq10}) or
continuous time in Eq. (\ref{eq4}) has stable fixed-point. Several known results are
listed, for example, in \cite{DynamicRNN}. An important case, as what our
main text was depending on is that, when the neural weights are symmetric,
the network converges no matter where it starts. In case convergence is
guaranteed with only one fixed-point, the corresponding dynamical system can
be regarded as dissipative with its fixed-point being globally
asymptotically stable. In this section, we denote ${\mathbf{y}}^{\ast }$ as
the position in the unit hypercube $\left[ {0,1} \right]^{n}$ that satisfies
the following.
\begin{equation}
\label{eq11}
{\mathbf{y}}^{\ast }=\sigma \left( {{\mathbf{Wy}}^{\ast }+{\mathbf{e}}}
\right)
\end{equation}
And unless specified otherwise, it is assumed that there is only one such
fixed-point. We further assume that there is a Lyapunov function $V$ of the
continuous system, such that for ${\mathbf{y}}\in \left[ {0,1} \right]^{n}$,
we have
\begin{equation}
\label{eq12}
\begin{array}{l}
 V\left( {{\mathbf{y}}} \right)=0,\text{\thinspace \thinspace iff\thinspace
}{\mathbf{y}}={\mathbf{y}}^{\ast } \\
 V\left( {{\mathbf{y}}} \right)>0,\text{\thinspace \thinspace if\thinspace
}{\mathbf{y}}\in \left[ {0,1} \right]^{n}\backslash \left\{
{{\mathbf{y}}^{\ast }} \right\} \\
 \left\langle {\frac{\partial V}{\partial
{\mathbf{y}}},-{\mathbf{y}}+{\mathbf{\lambda }}} \right\rangle \leqslant
0,\text{\thinspace \thinspace with\thinspace eq.\thinspace only\thinspace
at\thinspace }{\mathbf{y}}^{\ast } \\
 \end{array}
\end{equation}
Here ${\mathbf{\lambda }}$ is defined as in Eq. (\ref{eq9}). We focus on the case
when $\epsilon =1$ in which case Eq. (\ref{eq9}) can be rewritten as the following
\begin{equation}
\label{eq13}
\begin{array}{l}
 {\mathbf{y}}^{\left( {t+1} \right)}=\left( {1-a} \right)\cdot
{\mathbf{y}}^{\left( t \right)}+a\cdot {\mathbf{x}}^{\left( t \right)} \\
 x_{i}^{\left( t \right)} \sim \text{Bernoulli}\left( {\lambda_{i}^{\left(
t \right)} } \right) \\
 \end{array}
\end{equation}
Given the current position ${\mathbf{y}}^{\left( t \right)}\in \left[ {0,1}
\right]^{n}$ within a unit hypercube, ${\mathbf{\lambda }}^{\left( t
\right)}$ as defined in Eq. (\ref{eq9}) defines the tentative target where
${\mathbf{y}}^{\left( t \right)}$ will evolve towards, in case of a
deterministic system. While in Eq. (\ref{eq13}), the new position
${\mathbf{y}}^{\left( {t+1} \right)}$ is expressed as a convex combination
of the current position ${\mathbf{y}}^{\left( t \right)}$ and the target
${\mathbf{\lambda }}^{\left( t \right)}$, with combination ratios $1-a$ and
$a$ respectively. The stability of the stochastic system depends on two
factors: the value $a$ and where the fixed-point ${\mathbf{y}}^{\ast }$ is
located. When $a$ is closer to 1, ${\mathbf{y}}^{\left( {t+1} \right)}$ has
most of its part determined by the random sample rather than the last
position ${\mathbf{y}}^{\left( t \right)}$. Thus ${\mathbf{y}}$ along with
time $t$ evolves with large variation no matter where ${\mathbf{y}}^{\ast }$
is located. In particular, if $a>0.5$, we easily have
\begin{equation}
\label{eq14}
y_{i}^{\left( t \right)} \in \left[ {0,1-a} \right]\cup \left[ {a,1}
\right],\text{\thinspace }\forall i,t
\end{equation}
Thus no matter where ${\mathbf{y}}^{\ast }$ is located, there is a certain
region $\left( {1-a,a} \right)^{n}$ centered at the center of the hypercube
where ${\mathbf{y}}^{\left( t \right)}$ cannot reach at any time $t$. In
case ${\mathbf{y}}^{\ast }$ is inside that region, any trajectory will be
apart from ${\mathbf{y}}^{\ast }$ with at least a fixed distance, which
means that none of the stability results in \cite{StabSDE} holds true. In
another extreme, when $a$ is very small, since we have that $\left|
{y_{i}^{\left( {t+1} \right)} -y_{i}^{\left( t \right)} } \right|\leqslant
a$, the next position ${\mathbf{y}}^{\left( {t+1} \right)}$ will be within
the range of a small hypercube of length $2a$ centered at
${\mathbf{y}}^{\left( t \right)}$, and the Lyapunov function $V$ in this
small hypercube will be approximately linear. We denote the linearized
function $V$ passing through ${\mathbf{y}}^{\left( t \right)}$ as
$\Bar{{V}}$. Since the expectation of the next step is better in terms of
the Lyapunov function value, as the following
\begin{equation}
\label{eq15}
E\left[ {{\mathbf{y}}^{\left( {t+1} \right)}\vert {\mathbf{y}}^{\left( t
\right)}} \right]=\left( {1-a} \right)\cdot {\mathbf{y}}^{\left( t
\right)}+a\cdot {\mathbf{\lambda }}^{\left( t \right)}={\mathbf{y}}^{\left(
t \right)}-a\cdot \left( {{\mathbf{y}}^{\left( t \right)}-{\mathbf{\lambda
}}^{\left( t \right)}} \right)
\end{equation}
When $a$ is small, this approximately matches the evolution rule in the
deterministic system, which by assumption in Eq. (\ref{eq12}), guarantees to
decrease the Lyapunov function value. Thus we have
\begin{equation}
\label{eq16}
E\left[ {\Bar{{V}}\left( {{\mathbf{y}}^{\left( {t+1} \right)}} \right)\vert
{\mathbf{y}}^{\left( t \right)}} \right]<\Bar{{V}}\left(
{{\mathbf{y}}^{\left( t \right)}} \right)
\end{equation}
When the linear approximation is good (in case $a\to 0)$, the above
relationship with respect to $\Bar{{V}}$ also holds for $V$. This means that
the trajectory, although being stochastic, will almost deterministically
converge to the fixed-point and stay around there jumping randomly with
small perturbation. This phenomenon can be easily seen by numerical
simulation.

\begin{figure}[htbp]
\begin{center}
\begin{tabular}{cc}
\includegraphics[width=0.4\linewidth]{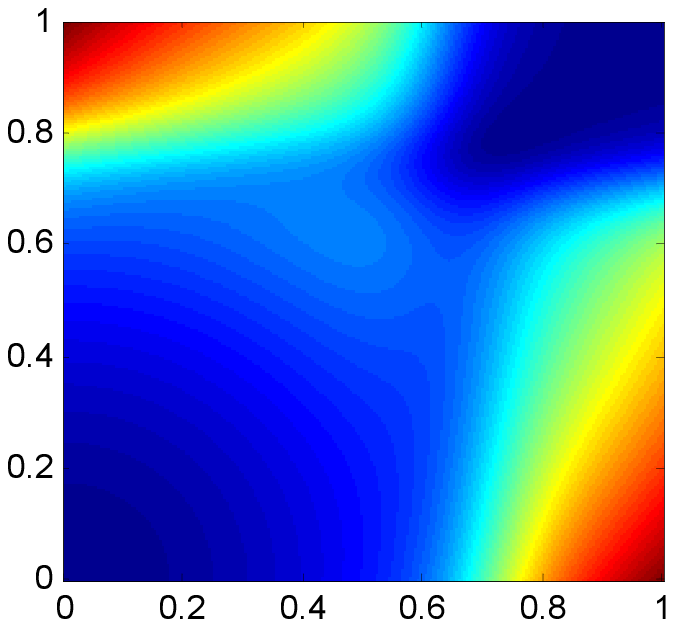} &
\includegraphics[width=0.4\linewidth]{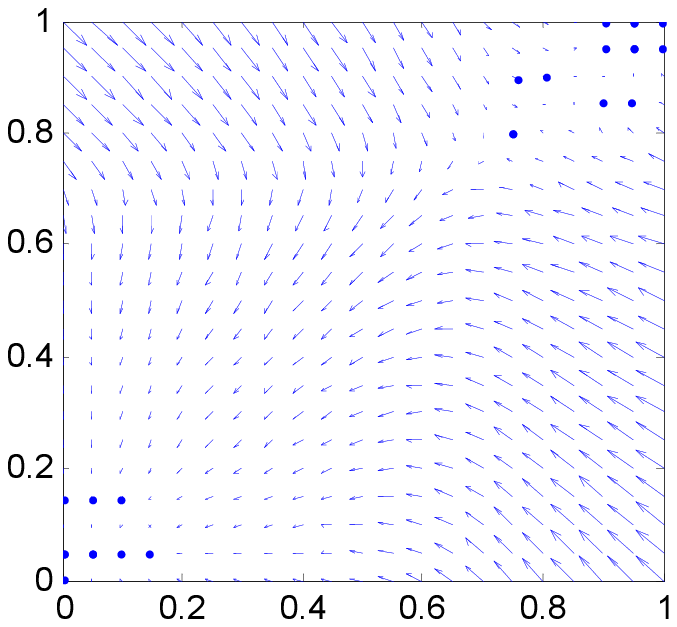} \\
(a)&
(b)
\end{tabular}
\end{center}
\caption{(a) Square of norm of r.h.s of Eq. (\ref{eq9}); (b) The vector field of
r.h.s of Eq. (\ref{eq9}).}\label{tab1}
\end{figure}

In particular, we found an interesting example. Suppose the parameters of
the network are as the following.
\begin{equation}
\label{eq17}
{\mathbf{W}}=\left[ {{\begin{array}{*{20}c}
 0 \hfill & {20} \hfill \\
 {15} \hfill & 0 \hfill \\
\end{array} }} \right],\text{\thinspace \thinspace }{\mathbf{e}}=\left[
{{\begin{array}{*{20}c}
 {-15} \hfill \\
 {-10} \hfill \\
\end{array} }} \right],\text{\thinspace \thinspace }a=0.5,\text{\thinspace
\thinspace }\epsilon =1
\end{equation}
Figure.~\ref{tab1} illustrates the deterministic dynamical
system defined by Eq. (\ref{eq17}). \tablename~\ref{tab1} (a) is
$\left\| {{\mathbf{y}}-{\mathbf{\lambda }}} \right\|^{2}$ plotted as color
for each ${\mathbf{y}}$, with red having high value and blue having low
value. \tablename~\ref{tab1} (b) is the vector field $-\left(
{{\mathbf{y}}-{\mathbf{\lambda }}} \right)$ showing how the deterministic
system will evolve. The dynamical system has two fixed points
${\mathbf{y}}_{1}^{\ast } =\left( {0.9922,0.9925} \right)$, and
${\mathbf{y}}_{2}^{\ast } =\left( {0.0031\text{e}-4,0.4540\text{e}-4}
\right)$. Starting from most of the positions in the unit hypercube, the
deterministic system will go to the stronger fixed-point
${\mathbf{y}}_{2}^{\ast } $. However, in the stochastic system as in Eq.
(\ref{eq13}), there are two situations. Most of the time the trajectory goes to
${\mathbf{y}}_{2}^{\ast } $ and stays there great stability. For very rare
cases the trajectory goes to ${\mathbf{y}}_{1}^{\ast } $ and stays there for
a while, but eventually jumps to ${\mathbf{y}}_{2}^{\ast } $. It could takes
long before such trajectory jumps back to ${\mathbf{y}}_{2}^{\ast } $, yet
in our more than 10000 test trials, no single trajectory stays in
${\mathbf{y}}_{1}^{\ast } $ forever.

The above example has very stable converged point, which is typical when the
deterministic system has one or more stable fixed-points that are very
strong (\textit{i.e.}, located near the vertices of the hypercube). In case
the fixed-point is not very strong, the situation is different. For example,
if when both $\left\| {{\mathbf{W}}} \right\|$ and $\left\| {{\mathbf{e}}}
\right\|$ are small, the following is also very small.
\begin{equation}
\label{eq18}
\left\| {{\mathbf{Wy}}^{\left( t \right)}+{\mathbf{e}}} \right\|\leqslant
\left\| {{\mathbf{W}}} \right\|\cdot \left\| {{\mathbf{y}}^{\left( t
\right)}} \right\|+\left\| {{\mathbf{e}}} \right\|
\end{equation}
In this case, ${\mathbf{\lambda }}^{\left( t \right)}$ will have its entries
close to 0.5 and thus each entry of ${\mathbf{x}}^{\left( t \right)}$ will
be very random. Making ${\mathbf{y}}^{\left( t \right)}$ evolves in changing
directions. The stability of ${\mathbf{y}}^{\left( t \right)}$ around
${\mathbf{y}}^{\ast }$ still depends on how large $a$ is. In case the
fixed-point ${\mathbf{y}}^{\ast }$ has different strength in each of its
entries, we found experimentally that those stronger entries have more
stable converged value, which is simply a combination of what we illustrated
in this section.

Experiments in our main text, especially Figure 3 and Figure 5 (a) (b),
demonstrated similar stability in this section.

\section{Open Questions}
Presented in the following are several questions related to Bayesian
inference by neural networks.

\subsection{Conditional Specification of Probability Models}
The connection between neuron models and Bayesian inference algorithms as
indicated in our main text is based on the observation that the neural
network structure, instead of representing the probability structure itself,
can be used to represent the ``information flow'' of a certain message
passing algorithm on a probability model. In this case the probability model
refers to the Boltzmann machine and the message-passing algorithm refers to
variational inference and Gibbs sampling. As mentioned in
\cite{CollapsedLDA}, many times variational inference has the mean-field
form of Gibbs sampling, thus these two algorithms may share the same
information flow structure. However, not necessarily all message-passing
algorithms can be directly encoded in neural networks. An important
requirement for doing so is that the message each neuron sends to its
downstream neurons must be the same. For example, belief-propagation (see
\cite{PRPearl}) and expectation propagation (see
\cite{DivergenceMeasure}) does not satisfy this requirement, so it's not
clear at this point whether these algorithms can be directly encoded in
neural networks.

Assuming an inference algorithm can indeed be encoded in neural networks,
the next question is then, given an arbitrary neural network, when does it
correspond to a valid joint probability model with the given inference
algorithm? This is important, for example, if we are capable of carrying out
both inference and learning on the neural networks without assuming the
joint probability model, we can simply carry out that algorithm (which
should be cheaper than actually learn probability semantics), and very
probably the learned model will exhibit correct probability semantics in the
limit, in the sense that the network structure and parameters under
learning, converge to those that do represent a joint probability model. For
example, the proof in \cite{Lebanon10SCL} can be used to show that if one
carry out the generalized backpropagation algorithm \cite{PinedaGBP} on
Boltzmann machines, and preserve the symmetric weights constraint, then
error-driven learning can yield models with correct probability semantics,
but what if the algorithm does not preserve symmetricity, such as the
Generalized Recirculation Algorithm \cite{GeneRec}.

In the context of this paper, the question is that, suppose the network
structure is to represent local conditional probability of a random
variable, when does the specification of a collection of local conditional
probability non-conflicting, and imply a uniquely determined joint
probability model. This is termed as the ``conditional specification
model'', see \cite{Arnold99CondSpec} for a reference. By the methodology in
that book, it can be easily seen that, assuming each neuron corresponds to
an individual binary random variable and the input set of neurons represent
the Markov blanket, the only way to let the neural network imply a joint
probability model which is compatible to all specified conditional
probabilities is that the weights must be symmetric. In another way, the
joint probability model that the conditionals of a neural network represent
has to be a Boltzmann machine. However, if we allow a many-to-one mapping,
\textit{i.e.}, different neurons can map to the same binary random variable,
the situation is different. As our main text implies, a neural network
without symmetric weights can still represent a Boltzmann machine if two
neurons specify the same binary random variable. Furthermore, if we allow
neurons to represent functions of random variables, or allow some hidden
neurons not representing any random variable, such that their purpose is
only for the supporting of the information flow, then much more
possibilities can happen. How do we even define this question rigorously?

\subsection{Universal Approximation by Distributions over Binaries}
Even though there are different ways of utilizing continuous variables in
neural networks, it may still be preferable to have all variables to be
encoded in binary. Since Boltzmann machines are universal approximator of
distributions over binary variables \cite{BengioRepresentationalPower}, the
natural question that follows is whether it is also a good approximation to
distributions over non-binary variables. If so, it would be sufficient for
representing knowledge when developing intelligent systems.

The question is twofold: (\ref{eq1}) Whether a continuous variable can be
efficiently encoded by a collection of binary variables, with the
approximation error being controllable. (\ref{eq2}) Whether distributions over the
continuous variables can be efficiently approximated by distributions over
the encoded binary variables.

When the continuous numbers are located on the unit interval, its binary
approximation is natural. We can simply take the binary digit representation
and truncate it to have the first set of digits. This approach may be
extended to domains where Fourier series functions can be well defined, in
which case the sign of the Fourier series function evaluations can be used
as a good binary approximation. In case the continuous domain does not have
Fourier series defined or such approximation is inefficient, we may be able
to look at the eigenfunctions of the Laplacian operator. As indicated in
\cite{BooleanFourier}, there are relationships between Fourier series and
eigenfunctions of Laplacian operator. This may provide a unified theoretical
framework for investigating on binary approximation.

{
\bibliography{BibTex_YuanlongShao}
\bibliographystyle{unsrt}
}